\documentclass[11pt]{article}
\usepackage{tikz}
\usetikzlibrary{angles, quotes}
\usepackage[margin=1in]{geometry}
\usepackage{amsmath}
\usepackage{iftex}
\ifPDFTeX
  \usepackage[T1]{fontenc}
  \usepackage{newtxtext}
  \usepackage{newtxmath}
  \usepackage[zerostyle=b,scaled=0.9]{newtxtt}
\else  
  \usepackage{fontspec}
  \defaultfontfeatures{Ligatures=TeX}  
  \usepackage{newtxtext}
  \usepackage{newtxmath}
\fi

\usepackage{graphicx}
\usepackage{longtable, booktabs, array}
\usepackage{calc}        
\usepackage{textcomp}    
\usepackage{xcolor}
\usepackage[font=small,labelfont=bf,labelsep=period]{caption}
\usepackage{fancyhdr}
\usepackage[numbers]{natbib}
\definecolor{linknavy}{HTML}{1A4B7C}
\definecolor{citepine}{HTML}{1F6E43}
\usepackage[colorlinks=true, linkcolor=linknavy, citecolor=citepine, urlcolor=linknavy]{hyperref}
\usepackage{microtype}
\usepackage{algorithm}
\usepackage{algpseudocode}
\usepackage{listings}   
\usepackage{needspace}  
\definecolor{promptrule}{gray}{0.55}
\DeclareCaptionFormat{promptcap}{\small\textbf{#1#2}#3}
\graphicspath{{.}}
\fancypagestyle{plain}{\fancyhf{}\fancyfoot[C]{\small\thepage}}
\providecommand{\tightlist}{\setlength{\itemsep}{0pt}\setlength{\parskip}{0pt}}

\title{\bfseries LLM-IDEA: Identifiability-Driven Experimental Agent for\\ Autonomous Discovery of Mechanistic World Models}
\author{Surya Shetty\\ Texas A\&M University\\ Polara Labs Inc. \and Ulisses Braga-Neto\\ Texas A\&M University}
\date{\today}

\begin{document}
\maketitle

\begin{abstract}
\small
Large language model agents are being increasingly deployed as autonomous scientists, designing experiments and inferring mechanistic world models with minimal human oversight. Yet identifiability is often overlooked: when a plateau is reached, the agent needs to know whether it is not yet capable enough or the model simply is not identifiable from the data, in which case no amount of further experimentation of the same kind can help. We propose the Identifiability-Driven Experimental Agent (LLM-IDEA) for closed-loop discovery with an identifiability engine that returns a three-way plateau verdict: capability limit, resolvable within the design class, or certified exhausted. On ODEBench, 60 of the 62 systems with free constants are identifiable at round 0; the RC circuit is certified exhausted for every experiment that protocol can run, and a harvesting model is resolvable by one added initial condition. The identifiability engine reproduces known verdicts on Lotka-Volterra, Van der Pol, Lorenz, and a pharmacokinetic model, where it recommends the intravenous arm pharmacologists use, and it ranks the depth scorer of our own benchmark last among four observation designs. On the DiscoverPhysics benchmark, it finds two public worlds whose explanation rubric rewards a distinction no legal experiment can make, and every model there with accurate trajectories failed the explanation grade (15 of 15, against 5 of 9 in identifiable worlds, $p = 0.012$). On the Alien Universe, a two-body testbed we propose in which a force law switches between a provably non-identifiable and an identifiable protocol, LLM-IDEA on the identifiable protocol reaches discovery depth at least three on 8/8 seeds versus 1/8 without it. An autonomous discovery agent can thus compute, rather than guess, whether a plateau calls for more search, a better experiment of the same kind, or a different kind of experiment.
\end{abstract}

\section{Introduction}\label{introduction}

The past two years have witnessed a rapid transition from large language models as passive question-answerers to active agents capable of planning, tool use, and closed-loop decision-making \citep{yao2023react, schick2023toolformer, wang2024survey}. This agentic turn has opened the door to autonomous scientific discovery, where AI systems design and execute experiments, analyze results, and iteratively refine hypotheses with minimal human intervention \citep{boiko2023autonomous, coscientist-nature-2026, sakana-ai-scientist}. A promising direction within this space is the recovery of {\em mechanistic world models}, a recently introduced framework for interpretable representations of the dynamical laws governing a system, which goes beyond pattern matching or predictive accuracy alone \citep{posner2026observation}.
Unlike black-box surrogates, mechanistic world models support extrapolation, intervention reasoning, and the transparent, falsifiable theorizing that lies at the heart of the scientific method \citep{cranmer2020discovering, schmidt2009distilling, battaglia2018relational, lake2017building, wang2023scientific}. 

It is against this backdrop of growing capability that a subtle but fundamental obstacle has gone largely unexamined: whether the experiments an autonomous agent can perform are even capable, in principle, of determining the parameters it seeks to identify.
An agent that learns a mechanistic world model will sometimes stop improving. That plateau may be encountered due to one of two possible causes. The first is capability: the agent is not yet good enough, and more search, a better proposer, or more compute would break through. But the second reason is identifiability: the parameter is not determined by the observation protocol that produced the data, and no further search will recover it. An agent that cannot tell these apart either keeps refining against a target its data cannot constrain or reports a confident world model that fits the sample but has no predictive power. It is worth pointing out that structural and practical identifiability analysis and Fisher-information experimental design have been used for decades in control, systems biology, and epidemiology to ask exactly this question \cite{ljung-glad-1994,atkinson2007optimum,raue-practical-identifiability,chowell2017fitting}.

To address this gap, we propose the Identifiability-Driven Experimental Agent (LLM-IDEA), which runs identifiability analysis on its own proposed world model and its own data, and then acts on the verdict. An LLM proposer, the Agentic Council, drafts hypotheses and candidate experiments. A deterministic engine with no language model in it decides what those experiments can reveal: it reads the rank of the sensitivity matrix, names each confounded parameter direction, ranks candidate designs by the rank they restore, and certifies when a direction is invisible to every design of the current kind. The certificate turns a plateau into one of three reports: keep searching (the data suffice), run a better experiment of the same kind, or change the kind of experiment. The idea that a sensitivity null space names what data cannot determine is old, and so is the idea that a symmetry shared by every admissible experiment cannot be broken by any of them. What is new here is making that decision inside an agent's loop, at the agent's own operating point, with a guarantee about when the stop is a proof. The main testbed is the Alien Universe, a two-body physics testbed we propose in which one force law can be switched between a provably non-identifiable observation protocol and an identifiable one. The engine is then pointed at systems we did not build, including the ODEBench suite a competing agent is evaluated on, and at the scoring metric of our own benchmark.
We summarize our contributions next.
\begin{enumerate}
\def\labelenumi{\arabic{enumi}.}
\item
  \textbf{Certified exhaustion and a three-way plateau triage, with soundness conditions.} A deterministic diagnostic computes the rank and conditioning of the sensitivity matrix of the proposed model on its own data and returns the confounded direction as a vector, with an exact symbolic check for rational and ODE forms. A certificate then decides whether that direction is a symmetry of the whole design class, which yields a three-way triage: capability limit, resolvable inside the class, or unresolvable by any design in the class. We show, as a consequence of Rothenberg's theorem, that the certificate is sound at regular points, give a counterexample at a singular point where a first-order certificate would stop the agent on an identifiable parameter, and show that any score computed from in-class predictions is blind to a certified direction (Section~\ref{certificate-theory}). Validation on systems we did not build is in Section~\ref{the-diagnostic-on-systems-we-did-not-build}; the live loop and the certificate are in Section~\ref{the-loop-closes-and-closes-by-proof}.
\item
  \textbf{The Alien Universe: a benchmark with a same-law observability switch.} This testbed holds one force law fixed and switches a single observation-protocol variable between a non-identifiable regime and an identifiable one, with the confounded direction known in advance. Under the identifiable protocol, the Agentic Council clears a cross-configuration gate that a single configuration cannot pass by construction, and that gate transfers to a second force-law regime. The comparison is a check that the switch works as designed rather than a capability measurement: the identifiable arm's task prompt also states that a per-body property exists, and runs share a distilled memory. A pre-registered two-by-two factorial experiment separates the protocol from the fitting engine (Section~\ref{observability-lift-and-second-regime-transfer}).
\item
  \textbf{The diagnostic turned on the benchmark's own metric.} The same rank-and-conditioning check, applied to the scorer that grades the agents, shows which sampling designs make the target exponent recoverable and which do not. The ranker orders the scorer's own design last among the candidates. Section~\ref{the-diagnostic-turned-on-the-benchmarks-own-metric} reports the designs and the bands.
\item
  \textbf{Validation beyond our testbed.} The diagnostic reproduces settled answers on textbook ODE systems and catches a published pharmacokinetic non-identifiability, where it recommends the design pharmacologists already use (Section~\ref{the-diagnostic-on-systems-we-did-not-build}). Swept over the 63 ODEBench systems under the LLM-ACES protocol, 62 of which have free constants, it finds 60 identifiable at round 0, one certified exhausted for the whole class of experiments LLM-ACES can run, and one resolvable inside that class (Section~\ref{the-diagnostic-on-odebench}). Applied to the rubrics of the DiscoverPhysics benchmark, it finds two worlds whose top explanation grade rewards a distinction no legal experiment can make, which is where models that fit the trajectories still fail (Section~\ref{discoverphysics-audit}). The unchanged loop also re-derives textbook integer-sequence identities blind, with a machine-checked generating-function certificate and a refused out-of-basis control (Section~\ref{transfer-to-integer-sequences-with-a-machine-checked-certificate}).
\end{enumerate}

No language model participates in validation; all scoring runs in numpy, sympy, and scipy behind an enforcement gate. Every live-agent measurement and the ODEBench, criterion-comparison, and DiscoverPhysics analyses were pre-registered; the textbook, pharmacokinetic, and integer-sequence analyses were exploratory, and we label them so (Section~\ref{protocol}). Each run operates under a fixed cost ceiling with an automatic halt.

The identifiability and experimental-design mathematics is prior work \cite{bellman-astrom-1970, cintron-arias-2009, stigter-molenaar-2015, ovchinnikov-2022-multiexperiment}; our contribution is the certified triage as an agent's stopping and design rule, its soundness conditions, and its use as an evaluation engine. The live loop that couples the engine to the proposer is a demonstration (five scored calls over a fixed two-move vocabulary, Section~\ref{the-loop-closes-and-closes-by-proof}), not a powered measurement. The Alien Universe lift carries confounds we state in Section~\ref{observability-lift-and-second-regime-transfer} and Section~\ref{limitations}. 


\section{Related Work}\label{related-work}

\subsection{Closed-loop LLM experiment design}\label{closed-loop-llm-experiment-design}

Independently of us, two recent works studied identifiability in an agentic system consisting of a language-model proposer and an experiment-selection step for learning mechanistic world models.
The Model Discovery Agent \cite{murphy-mda-2026} takes the Bayesian route: it keeps a sequential Monte Carlo posterior over models and parameters and picks the next design by value of information, while LLM-ACES \cite{llm-aces-2026} takes a deterministic route like ours: it picks the next trajectory by pairwise disagreement among fitted candidates and names the resulting identifiability gap, meaning observationally indistinguishable equation structures; LLM-AutoSciLab is a related framework from the same group~\cite{llm-autoscilab-2026}. Both methods treat identifiability as something the next experiment resolves. Neither computes a sensitivity or Fisher rank for the agent's current model, names the confounded parameter direction as a vector, or returns, as an output of its loop, the verdict ``no experiment of this kind can help,'' which separates a merely bad design from a wrong class of designs. The MDA authors do observe such floors in their own analysis \cite{murphy-mda-2026}; our point is that the agent can compute and act on the verdict itself.
CARTOGRAPH \cite{shah-kartik-2026-cartograph} is the closest system on the engine side. It stacks disagreement Jacobians into a matrix whose kernel is the unresolved parameter subspace, steers experiments by (A-optimal) Fisher information on that subspace, and refuses identifications that residual analysis contradicts, with pharmacokinetic and autonomous-materials audits. It has no language-model proposer in its main loop, and it does not decide whether a direction is invisible to every design in a class; that class-level certificate, its soundness conditions, and the three-way triage it produces are what we add. Its A-optimal ranking is more principled than our rank-then-conditioning heuristic (Section~\ref{the-design-ranker}). 



\subsection{Identifiability and Observability}\label{identifiability-observability-and-experimental-design}

Structural identifiability, whether parameters are uniquely determined by perfect noise-free data, was defined by Bellman and Åström \cite{bellman-astrom-1970} and treated differential-algebraically by Ljung and Glad \cite{ljung-glad-1994}. Reviews are provided by  \cite{chis-banga-balsa-canto-2011} and \cite{wieland-2021}, while specific frameworks include STRIKE-GOLDD \cite{villaverde-2016-strike-goldd}, SIAN \cite{sian}, and StructuralIdentifiability.jl \cite{structidjl}. The work in \cite{villaverde-2019-observability} unifies identifiability with observability, rooted in Kalman's observability theory \cite{kalman-1960} and in \cite{hermann-krener-1977}. Practical identifiability and recoverability from finite noisy data uses sensitivity and Fisher information matrices \cite{raue-practical-identifiability}, whose ``sloppy'' spectrum is characterized in \cite{gutenkunst-2007,transtrum-2015}. Our rank-then-Fisher procedure follows \cite{cintron-arias-2009}; reading estimable combinations from the null vectors of the derivative matrix goes back to \cite{catchpole-morgan-1997,cole-morgan-titterington-2010}, and numeric sensitivity SVD followed by symbolic confirmation is \cite{stigter-molenaar-2015}. The same concern recurs in epidemiological calibration \cite{epi-identifiability}. Non-identifiability is equivalently a continuous symmetry of the parameter-to-output map, found and broken symbolically by Lie-symmetry methods \cite{yates-evans-chappell-2009,merkt-timmer-kaschek-2015,massonis-villaverde-2020}; which outputs or inputs break a symmetry is studied as minimal output sets \cite{anguelova-2012-minimal-outputs} and input-dependent identifiability \cite{villaverde-2019-input}, and what any number of experiments with generic initial conditions can identify is characterized in \cite{ovchinnikov-2022-multiexperiment}. Our certificate is the operational, at-the-operating-point version of that class-level question, run inside an agent's loop; its symbolic route checks only symmetries that act on parameters, not ones that also transform states. 
The frameworks in \cite{liu-maini-baker-2025} and \cite{norden-2025} bring identifiability into model discovery and learning with partially observed dynamics. The two-parameter Lotka-Volterra form in our registry (Section~\ref{the-diagnostic-on-systems-we-did-not-build}) is the pysindy example \cite{brunton-2016-sindy}.

\subsection{End-to-End AI scientists}\label{end-to-end-ai-scientists}

First-generation systems introduced the concept of LLM agents for scientific discovery. Some examples of such systems include the Sakana AI Scientist \cite{sakana-ai-scientist,sakana-ai-scientist-nature-2026}, Google's AI co-scientist \cite{coscientist-nature-2026}, Kosmos \cite{kosmos-2025}, AlphaEvolve \cite{alphaevolve-2025}, ATHENA \cite{toscano2025athena,toscano2026graft} and FunSearch \cite{funsearch}. These frameworks share a generate-and-fit pattern, and none, as far as we can tell, tests whether its plateau comes from a weak proposer or from an answer absent from the data. 
The AI Scientific Community programme \cite{braga-neto-swarm-2026,loo-braga-neto-2026} is an extension of these first-generation systems to swarms of agentic virtual laboratories, which simulate the evolution of a scientific community around a specific problem.

\subsection{Discovery Benchmarks}\label{discovery-benchmarks}

NewtonBench \cite{newtonbench} tests law rediscovery across many physics scenarios. Gravity-Bench \cite{gravity-bench} poses gravitational inference with a budgeted observation interface. DiscoverPhysics \cite{discoverphysics} and PhysGym \cite{physgym} extend these. LLM-SRBench \cite{llm-srbench} and SR-Scientist \cite{sr-scientist} benchmark LLM symbolic regression against ground-truth equations, while AstaBench \cite{astabench} evaluates research agents more broadly. BoxingGym \cite{boxinggym-2025}, SciGym \cite{scigym-2025}, Stargazer \cite{stargazer-2026}, and MaD Physics \cite{mad-physics-2026} add interactive environments spanning experimental design, systems biology, astrophysics, and physical information seeking. These supply the answer key and score success rate. By contrast, our proposed 
testbed holds one law fixed and switches a single observation-protocol variable between a provably non-identifiable and an identifiable regime, with the confounded direction named in advance. Gravity-Bench varies observation density, not identifiability. ActiveSciBench \cite{llm-autoscilab-2026} grades tasks by identifiability strength but does not expose it as a switch. NewtonBench's identifiability argument establishes solvability under oracle access and asymptotic identifiability given a rich enough sample. 

\section{Methodology}\label{methodology}

\subsection{The Agentic Council}\label{the-loop}

The proposer is the Agentic Council (hereafter the Council), a five-role multi-agent system in which one role (the synthesizer) drafts hypotheses and candidate designs from the other four roles' contributions. The Council's roles share a knowledge graph into which one role also distills judging criteria from past proposals; those criteria are injected into later prompts (Section~\ref{limitations} discusses what that sharing does to seed independence). The framework also contains a three-role deliberation court (proponent, critic, judge) for contested hypotheses, but the court issued no model call in any run reported here, and no result depends on it. Neither touches the identifiability diagnostic. On a plateau, if the hypothesis is deemed non-identifiable by the identifiability engine, the Council proposes candidate designs, the ranker selects one, the experiment is run, and the diagnostic re-checks. The loop terminates as \texttt{solved} when re-diagnosis returns full rank, as \texttt{exhausted\_by\_certificate} when the certificate of Section~\ref{certified-exhaustion-and-plateau-triage} proves no design of the current kind can help, or as \texttt{exhausted} when no proposed experiment breaks the degeneracy within the round budget. 

\subsection{The Identifiability Engine}\label{the-identifiability-engine}

The identifiability engine is a deterministic procedure that decides whether an open scientific question is answerable from the data in hand. When it is not, it names the confounded direction, decides whether any experiment of the current kind could resolve it, and if so selects the measurement that would. It has four parts: a diagnostic, a design ranker, a certificate of exhaustion, and a loop coupling them to an autonomous proposer. No language model participates in any of the four. The diagnostics are objectively computed in \texttt{numpy}, \texttt{sympy} and \texttt{scipy}, and their correctness does not depend on a model's judgment.

\subsection{The Diagnostic}\label{the-diagnostic}

Given a model class and an observation protocol, the diagnostic decides whether the parameters are recoverable from the observations and, if not, names the degeneracy. For models with rational or ordinary-differential-equation structure we use an exact symbolic structural test \cite{bellman-astrom-1970}. For the general case, where a hypothesis may be black-box or non-ODE, we use practical identifiability via the sensitivity matrix and its Fisher information. That route assumes only that predictions can be evaluated as a function of the parameters. Structural-identifiability tools assume rational-function dynamics \cite{ljung-glad-1994,chis-banga-balsa-canto-2011,wieland-2021}, and an autonomous agent's hypotheses often do not, so practical identifiability is the general fallback.

\def\bth{\boldsymbol{\theta}}
Let $\mathbf{y} \in \mathbb{R}^d$
be the noisy vector of observations that the design produces corresponding to the parameter vector $\boldsymbol{\theta} \in \mathbb{R}^p$. We write the model as
\begin{equation}
\mathbf{y} = \mathbf{f}(\bth) + \boldsymbol{\varepsilon},\qquad \boldsymbol{\varepsilon} \sim \mathcal{N}(\mathbf{0},\sigma^2 I),
\end{equation}
where $\mathbf{f}: \mathbb{R}^p \rightarrow \mathbb{R}^d$ is the mechanistic model. In order to test the identifiability of a model specified by $\bth$, we use the \emph{local sensitivity matrix} $S$, which is the Jacobian of $\mathbf{f}$ computed at $\bth$:
\begin{equation}
S(\bth)_{ij} \,=\, \frac{\partial f_i(\bth)}{\partial \theta_j}\,,
\end{equation}
where the partial derivatives are typically intractable to compute analytically and must be approximated numerically. Here, we use central finite differences, or the forward sensitivity equations when the model is an ODE (Section~\ref{the-diagnostic-on-odebench}).
Based on the local sensitivity matrix, there are three verdicts \cite{Rothenberg1971}:
\begin{enumerate}
\item[1.] If $S(\bth)$ is full-rank and its condition number is at most $10^8$, the model is locally identifiable at $\bth$.
\item[2.] If $\operatorname{rank}(S(\bth)) < p$, the model is locally non-identifiable at $\bth$, and the null space of $S(\bth)$ gives the non-identifiable combinations of values in $\bth$. Rothenberg's equivalence requires $\bth$ to be a regular point, where the rank is constant nearby; at a singular point a rank drop need not mean non-identifiability (Section~\ref{certificate-theory}).
\item[3.] If $S(\bth)$ is full-rank but its condition number exceeds $10^8$, the model is practically non-identifiable~\cite{gutenkunst-2007,transtrum-2015}.
\end{enumerate}
Reading the null space of the sensitivity matrix as the unidentifiable combinations is classical \cite{catchpole-morgan-1997,cole-morgan-titterington-2010,stigter-molenaar-2015}; what we add is its use as a live stopping and design signal for a language-model proposer.

Two observations are important. First, the noise level $\sigma$ changes neither the rank nor the condition number of $S$, and its value matters only if the Fisher information matrix or its eigenvalues are required. Second, the analysis is local at the value of $\bth$. When global behavior is in question, probing several operating points is the remedy, and Section~\ref{the-diagnostic-on-systems-we-did-not-build} shows a published case where the local check misses a solution.


\subsection{The Design Ranker}\label{the-design-ranker}

Given a set of candidate observation designs, each a different protocol over which inputs, configurations, or times are measured, the ranker scores every candidate by how well it identifies the target parameters at the shared operating point. It recommends the disambiguating one. Rank dominates, with each recovered parameter direction worth a full point, and conditioning is a sub-unit tie-break. This is a heuristic built on the same Fisher matrix that classical optimal experimental design uses \cite{fedorov-1972,pukelsheim-2006}. It is neither D-optimality nor E-optimality, nor a Bayesian expected-information-gain criterion \cite{chaloner-verdinelli-1995,huan-marzouk-2013}. Section~\ref{criterion-head-to-head} compares it head to head with those criteria on identical candidates: it restores rank as reliably as any of them, and A-optimal design chooses more precise experiments. A candidate whose \texttt{predict} closure fails to evaluate (a malformed proposal) is scored worst and never recommended, but does not crash the ranking.

In the Alien Universe case the ranker reconstructs the fix that closed the original discovery ceiling. At a single configuration the charge product is confounded with the force amplitude. Given candidates that include configurations at distinct products, the ranker selects the multi-product family, because only that design restores rank. A fix once found by hand is now the output of a domain-agnostic primitive.

\subsection{Certified exhaustion and plateau triage}\label{certified-exhaustion-and-plateau-triage}

An \texttt{exhausted} outcome after a fixed number of rounds says only that the proposer did not find a fix, not whether one exists. A weak proposer and a design class in which no experiment could ever help look the same. The certificate separates them.

The certificate asks whether the confounded direction is a symmetry of the entire design class rather than of the single diagnosed design. The symbolic route takes the model as a \texttt{sympy} expression and fixes \texttt{theta} at the operating point as exact rationals. It keeps every observable and design control (time, radius, dose, initial condition) symbolic, snaps the numeric null vector to small rationals, and checks whether the directional derivative of the prediction along that vector simplifies to literal zero. If so, the direction is invisible to every design in the class, a proof at that operating point. The numeric route stacks sensitivity matrices from many sampled designs of the class and tests whether the direction is still null across the stack, with a normalized response at or below \texttt{1e-7}. It needs no closed form, but it is evidence about the sampled designs only, and Section~\ref{limitations} says so.

A plateau is then triaged three ways. If the current design is already identifiable, the plateau is a capability limit (\texttt{capability}): the data suffice and the agent has not yet found the law. If it is non-identifiable and at least one confounded direction is not certified, the plateau is resolvable inside the class (\texttt{design\_within\_class}), and the ranker is the right tool. If every confounded direction is certified, no experiment of this kind can resolve it (\texttt{change\_class}); the right move is to change the kind of experiment or report the identifiable combination. The loop runs the certificate at round zero. On \texttt{change\_class} it terminates immediately with outcome \texttt{exhausted\_by\_certificate}, spending no proposer calls.

Section~\ref{the-loop-closes-and-closes-by-proof} exercises the certificate on three plateaus. The first is the Alien Universe single-configuration plateau, where the charge swap is certified for every design and the scaling of $\beta$ against the charges is certified for every class in which no charge is known, a family included. The second is the pharmacokinetic model of Section~\ref{the-diagnostic-on-systems-we-did-not-build}, where an oral-only class is certified exhausted. The third is the integer-sequence window of Section~\ref{transfer-to-integer-sequences-with-a-machine-checked-certificate}, where the collapse is not certified and the ranker is the right tool. This verdict is the concrete difference from the closed-loop systems of Section~\ref{closed-loop-llm-experiment-design}.

\subsection{When the certificate is a proof}\label{certificate-theory}

A design class $\mathcal{C}$ is the set of experiments the agent may run: each design $d \in \mathcal{C}$ yields predictions $f(d;\bth)$, and in the symbolic route the free variables of the model expression (times, radii, doses, initial states) range over $\mathcal{C}$. The \emph{class sensitivity} at $\bth$ is 
\begin{equation}
\mathcal{S}_{\mathcal{C}}(\bth) \,=\, \operatorname{span}\{\nabla_{\bth} f(d;\bth) : d \in \mathcal{C}\} \subseteq \mathbb{R}^p.
\end{equation}
The dimensionality $r_{\mathcal{C}}(\bth)$ of $\mathcal{S}_{\mathcal{C}}(\bth)$ is the \emph{class rank}, and its orthogonal complement $\mathcal{N}_{\mathcal{C}}(\bth)$ is the set of directions that no design in the class sees to first order. The symbolic route of Section~\ref{certified-exhaustion-and-plateau-triage} tests $u \in \mathcal{N}_{\mathcal{C}}(\bth_0)$ exactly, using the fact that the directional derivative $u \cdot \nabla_{\bth} f(d;\bth_0)$ must reduce to zero with every free variable.

However, that test alone does not prove non-identifiability. For example, take $f(x;b) = b^3 x^2$ with design $x>0$, at $b = 0$. Every design has $\partial f/\partial b = 3b^2x^2 = 0$ there, so the direction $b$ passes the first-order test for the whole class, yet $b$ is identifiable, since $b \mapsto b^3 x^2$ is injective for any design $x > 0$. The class rank is 0 at $b = 0$ and 1 at every nearby point. A first-order certificate would tell the agent to stop on a parameter a single experiment would pin down. The missing condition is the regularity that Rothenberg's theorem already requires for a single design \cite{Rothenberg1971}, applied to the whole class.

\vspace{1ex}
\noindent
\textbf{Proposition 1 (soundness of \texttt{change\_class}).} {\em Let each $f(d;\cdot)$ be continuously differentiable, and let the class rank be constant, equal to $r$, on a neighborhood of $\bth_0$. Then every $u \in \mathcal{N}_{\mathcal{C}}(\bth_0)$ is tangent at $\bth_0$ to a $(p-r)$-dimensional submanifold $M$ through $\bth_0$ on which $f(d;\bth) = f(d;\bth_0)$ for every $d \in \mathcal{C}$. No design in the class, and no finite set of designs in it, distinguishes $\bth_0$ from the points of $M$.}

\smallskip\noindent
This is Rothenberg's Theorem~1 \cite{Rothenberg1971} applied to the joint model that observes every design in $\mathcal{C}$. The class rank is attained by finitely many designs $d_1,\ldots,d_r$, so the stacked map $\bth \mapsto (f(d_i;\bth))_{i \le r}$ has constant rank $r$ near $\bth_0$, and the constant-rank argument of \cite{Rothenberg1971} gives the submanifold $M$; since adding any other design of the class leaves the rank at $r$, every $f(d;\cdot)$ with $d \in \mathcal{C}$ is constant on $M$. The same rank condition underlies the parameter-redundancy results for exhaustive summaries \cite{catchpole-morgan-1997,cole-morgan-titterington-2010}.

\vspace{1ex}
\textbf{What the other two verdicts guarantee.} If $u \notin \mathcal{N}_{\mathcal{C}}(\bth_0)$, some single design in $\mathcal{C}$ has a sensitivity row not orthogonal to $u$, and at most $r_{\mathcal{C}}(\bth_0)$ designs from the class reach the class rank. No set of designs from the class exceeds it. So \texttt{design\_within\_class} is always resolvable to first order by a finite in-class design, and the most any in-class experiment can deliver is rank $r_{\mathcal{C}}(\bth_0)$: once some directions are certified, full rank is unreachable inside the class and the correct report is the identifiable combinations. (Immediate from the definition of the span.) The implementation therefore returns \texttt{solved\_up\_to\_class\_symmetry} when, after at least one design round, every remaining confounded direction is certified: the experiments have reached the class rank, and what remains can only be reported as identifiable combinations or broken by a change of class.

\textbf{In-class metrics are blind to certified directions.} Under the hypotheses of Proposition~1, any score that depends on a fitted parameter $\hat\bth$ only through its predictions at designs in $\mathcal{C}$ (a held-out trajectory error on in-class test designs, for example) takes the same value at every point of $M$. Such a score can neither reward nor penalize the agent along a certified direction. (Immediate from Proposition~1.) This is why no held-out error on in-class trajectories can see the certified RC-circuit direction of Section~\ref{the-diagnostic-on-odebench} (the LLM-ACES symbolic metric misses it for a simpler reason: it strips constants), and why a test of whether non-identifiability explains where agents fail must use out-of-class or parameter-level metrics.

\textbf{Implementation.} Both certificate routes now check regularity before certifying. The symbolic route evaluates the parameter gradients at exact rational $\bth_0$ and at random rational values of the free variables in 50-digit arithmetic, computes the class rank there and at three nearby rational parameter points, and refuses when the rank rises off $\bth_0$. For analytic models the parameters where the class rank falls below its generic value form a proper analytic subset, so a singular $\bth_0$ is detected with probability one. The numeric route applies the same comparison to its stacked finite-difference sensitivity matrix. Both routes certified the counterexample above before this check was added, and both refuse it now. Every certificate reported in this paper passes at a verified regular point: the Alien Universe swap and scaling directions (class rank 1 at $\bth_0$ and nearby), the oral pharmacokinetic $(V, F)$ direction (3 and 3), and the RC circuit (2 and 2).

\subsection{Algorithm and usage}\label{algorithm-1-and-usage}

\begin{algorithm}[tb]
\caption{Identifiability-driven loop with certified exhaustion}
\label{alg:identifiability-loop}
\begin{algorithmic}[1]
  \Require model \texttt{predict}($\theta,d$), operating point $\theta$, current design $d_0$, design class $\mathcal{C}$, proposer $P$, maximum rounds $R$
  \Ensure outcome in \{\texttt{capability\_plateau}, \texttt{solved}, \texttt{solved\_up\_to\_class\_symmetry}, \texttt{exhausted}, \texttt{exhausted\_by\_certificate}\} and a report
  \State $v \gets \operatorname{diagnose}(\texttt{predict}, \theta, d_0)$ \Comment{rank, condition, and null directions}
  \If{$v.\texttt{rank} = n_{\mathrm{params}}$}
    \State \Return $(\texttt{capability\_plateau},\theta)$ \Comment{the data suffice; keep searching}
  \EndIf
  \ForAll{$u \in v.\texttt{null}$}
    \State $\mathrm{cert}[u] \gets \operatorname{Certify}(u,\theta,\mathcal{C})$ \Comment{Algorithm~\ref{alg:certify}}
  \EndFor
  \If{all $\mathrm{cert}[u]$ are certified}
    \State \Return $(\texttt{exhausted\_by\_certificate},\text{identifiable combinations of }\theta)$
  \EndIf
  \For{$r \gets 1,\ldots,R$}
    \State $\mathcal{D} \gets P.\texttt{propose\_designs}(v)$ \Comment{language model proposes; never judges}
    \State $d^\star \gets \operatorname{rank\text{-}designs}(\mathcal{D},\theta)$ \Comment{rank first, conditioning as tie-break}
    \State $\mathrm{data} \gets \operatorname{run\text{-}experiment}(d^\star)$
    \State $v \gets \operatorname{diagnose}(\texttt{predict},\theta,d^\star)$ \Comment{re-diagnose the chosen design}
    \If{$v.\texttt{rank} = n_{\mathrm{params}}$}
      \State \Return $(\texttt{solved},\operatorname{fit}(\theta,\mathrm{data}))$
    \EndIf
    \If{every $u \in v.\texttt{null}$ is certified}
      \State \Return $(\texttt{solved\_up\_to\_class\_symmetry},\text{identifiable combinations})$ \Comment{class rank reached}
    \EndIf
  \EndFor
  \State \Return $(\texttt{exhausted},\text{identifiable combinations of }\theta)$
\end{algorithmic}
\end{algorithm}

Algorithm~\ref{alg:certify} expands the \texttt{Certify} step. It answers one question about one confounded direction $u$: is $u$ invisible to \emph{every} design in the class $\mathcal{C}$, and not only to the design already run? The symbolic route answers with a proof when the model is available in closed form; the numeric route answers with evidence over sampled designs when it is not. Both routes perform two checks. The first is the first-order condition: the prediction does not move along $u$ for any design. The second is the regularity condition of Section~\ref{certificate-theory}: the class rank at $\theta$ equals the class rank at nearby parameter values. Without the second check the certificate can stop the agent on a parameter that two more measurements would identify (the counterexample of Section~\ref{certificate-theory}).

\begin{algorithm}[tb]
\caption{Certify: is a confounded direction a symmetry of the whole design class?}
\label{alg:certify}
\begin{algorithmic}[1]
  \Require confounded direction $u$, operating point $\theta$, design class $\mathcal{C}$; either a closed-form prediction $f(x;\theta)$ whose free variables $x$ (times, radii, doses, initial states) range over $\mathcal{C}$, or a sampler of designs from $\mathcal{C}$
  \Ensure \texttt{certified} or \texttt{not certified}, with the reason
  \Statex \textbf{Symbolic route} (closed form available; a proof)
  \State $\hat u \gets$ $u$ rounded to small rationals; $\theta$ written as exact rationals
  \State $g(x) \gets \sum_j \hat u_j\, \partial f(x;\theta)/\partial\theta_j$ \Comment{directional derivative, $x$ kept symbolic}
  \If{$g(x)$ does not simplify to $0$}
    \State \Return \texttt{not certified}: some design in $\mathcal{C}$ responds along $u$ \Comment{ask the ranker for it}
  \EndIf
  \State $r_0 \gets \operatorname{ClassRank}(\theta)$
  \For{$k \gets 1, 2, 3$}
    \State $\theta_k \gets \theta + $ small random rational perturbation; \quad $r_k \gets \operatorname{ClassRank}(\theta_k)$
  \EndFor
  \If{$r_0 < \max_k r_k$} \Return \texttt{not certified}: $\theta$ is a singular point \Comment{Section~\ref{certificate-theory}}
  \EndIf
  \State \Return \texttt{certified}: by Proposition~1, a $(p - r_0)$-dimensional set of parameters through $\theta$ is indistinguishable by every design in $\mathcal{C}$
  \Statex \textbf{Numeric route} (black-box model; evidence over sampled designs)
  \State draw designs $d_1,\ldots,d_N$ from $\mathcal{C}$; $S \gets$ the stacked sensitivity matrices $[S_{d_1}; \ldots; S_{d_N}]$ at $\theta$
  \If{$\lVert S \hat u\rVert / \lVert S\rVert > \tau$} \Return \texttt{not certified} \Comment{$\tau = 10^{-7}$ by default}
  \EndIf
  \If{$\operatorname{rank}(S)$ at $\theta$ $<$ $\operatorname{rank}(S)$ at some of three nearby $\theta_k$} \Return \texttt{not certified} \Comment{singular point}
  \EndIf
  \State \Return \texttt{certified} over the sampled designs \Comment{evidence, not proof}
  \Statex
  \Function{ClassRank}{$\vartheta$} \Comment{dimension of $\operatorname{span}\{\nabla_\theta f(x;\vartheta) : x \in \mathcal{C}\}$}
    \State draw $p + 6$ random rational points $x_i$ in the class
    \State \Return the rank of the matrix with rows $\nabla_\theta f(x_i;\vartheta)$, computed in 50-digit arithmetic with a $10^{-30}$ relative floor
  \EndFunction
\end{algorithmic}
\end{algorithm}





\section{The Alien Universe Testbed}\label{the-alien-universe-testbed}

Here we describe our proposed testbed consisting of a two-dimensional two-body problem whose governing law contains a hidden per-body property. The physics of this problem are invented and thus the name ``Alien Universe''.
Under one observation protocol the hidden degree of freedom is provably non-identifiable; under another it is identifiable. The switch is a deterministic property of the data that we control and is computed before any agent runs. We know of no other agentic-discovery benchmark that switches the identifiability of a single fixed law.

\subsection{The Dynamical System}\label{the-law-as-simulated}


\begin{figure}[t!]
\begin{center}
\begin{tikzpicture}

  \def\angle{35}
  \def\radius{5}

  \coordinate (O) at (0,0);

  \coordinate (M1) at (0,0);

  \coordinate (M2) at (\angle:\radius);

  \draw[dashed, gray] (M1) -- ++(3,0) coordinate (Href);

  \draw[thick] (M1) -- (M2) node[midway, above left, font=\large] {$r$};

  \pic[
    draw=black,
    thick,
    angle radius=1.5cm,
    angle eccentricity=1.3,
    "$\theta$"
  ] {angle = Href--M1--M2};

  \filldraw[fill=blue!40, draw=blue!70!black, thick] (M1) circle (0.45cm);
  \node[below left=6pt, font=\large] at (M1) {$m_1,s_1$};

  \filldraw[fill=red!40, draw=red!70!black, thick] (M2) circle (0.25cm);
  \node[above right=4pt, font=\large] at (M2) {$m_2,s_2$};
\end{tikzpicture}
\end{center}
\caption{The two-body problem in the Alien Universe. The bodies have masses and charges $m_1,s_1$ and $m_2,s_2$.}
\label{Fig-twobody}
\end{figure}
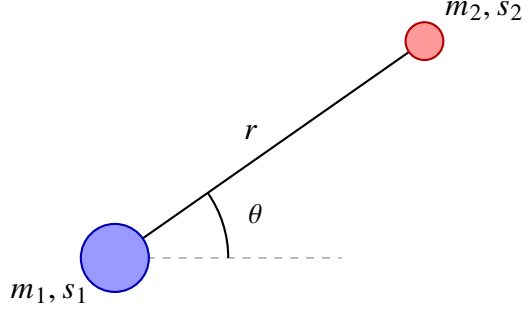

Consider two bodies of masses and charges $m_1,s_1$ and $m_2,s_2$ in a plane, separated by a scalar distance $r$ and oriented at an angle $\theta$. Please see Fig~\ref{Fig-twobody} for an illustration. In the Alien Universe, the gravitational force between the bodies is an inverse-square attraction force with a short-range correction,
\begin{equation}
F_g(r) \,=\, G m_1 m_2 \left(\frac{1}{r^2} + \frac{\alpha}{r^{3.5}}\right)\,,
\end{equation}
where $\alpha > 0$ is a tunable coupling, while the electric force is a non-polynomial repulsion force,
\begin{equation}
F_s(r,s_1,s_2) \,=\, \beta s_1 s_2 \frac{1-\cos(\gamma r)}{r^2}\,,
\end{equation}
where $\gamma>0$ is a fixed inverse-length scale parameter.
The dynamics of the system are governed by the Hamiltonian ODEs:
\begin{equation}
\begin{aligned}
&\dot r\,=\,\frac{p_r}{\mu}\,,\quad\quad
\dot\theta\,=\,\frac{p_\theta}{\mu_\perp r^2},\\[1ex]
&\dot p_r\,=\,\frac{p_\theta^2}{\mu_\perp r^3}-F_g(r)+F_s(r,s_1,s_2)\,,\quad\quad
\dot p_\theta\,=\,0,
\label{eq:alien-ode}
\end{aligned}
\end{equation}
where $p_r$ and $p_\theta$ are the radial and angular momenta, 
\begin{equation}
\mu\,=\,\frac{m_1m_2}{m_1+m_2}\,.
\label{eq:reduced-mass}
\end{equation}
is the reduced mass from standard central-force theory \cite{goldstein2002classical}, and
\begin{equation}
\mu_\perp \,=\, \mu + \kappa s_1 s_2,
\label{eq:tangential-mass}
\end{equation}
is a tangential inertia term that affects motion perpendicular to the axis joining the bodies, with $\kappa > 0$ a coupling coefficient.

We integrate the ODE system to obtain the trajectories $r(t)$ and $\theta(t)$, which are converted back to the two Cartesian coordinates $x(t)$ and $y(t)$, with the velocity trajectories $\dot x(t)$ and $\dot y(t)$ shown to the agents. At the tuned operating point of the first regime (the second regime of Section~\ref{observability-lift-and-second-regime-transfer} changes the coupling and the correction exponent), $G = 1$, $\alpha = 0.05$, $\beta = 0.02$, $\gamma = 0.7$, $\kappa = 0.3$, and $m_1 = m_2 = 1$; the integrator is the fixed-step implicit midpoint rule with step $0.005$ over $1000$ steps, and the radial force is obtained from the potential by a fourth-order finite difference.

 The inertia coupling is velocity-dependent, invisible to any radial-force fit; the charge surfaces only as a systematic angular discrepancy correlated with the charge product.
Because the charges enter only through \(s_1 s_2\), equal-product configurations integrate to bit-identical trajectories, the basis of the switch in Section~\ref{the-observability-switch}. The \(r^{-2}\) and \(r^{-3.5}\) terms demand different time-scaling exponents, so the law is not scale-invariant and does not collapse to polynomial or power-law regression. The law exists only in code we wrote, so no step of recovering it can be retrieved from a training corpus.  Its mechanics are nevertheless standard: the center-of-mass reduction, polar Hamiltonian, and angular-momentum conservation follow textbook central-force mechanics \cite{goldstein2002classical}, and the fixed-step implicit-midpoint method is a standard symplectic integrator \cite{hairer2006geometric}.  What is bespoke is the hidden force \(F_s\) and the charge-dependent tangential inertia of Eq.~\ref{eq:tangential-mass}; we did not adapt those terms from a prior physical model.

\subsection{The depth ladder}\label{the-depth-ladder}

We score how far an agent gets along an ordered ladder:
\begin{itemize}
\tightlist
\item
  \textbf{L1, dominant force law.} There is an inverse-square attractive force;
\item
  \textbf{L2, correction term.} Residuals against the base force are systematic; there is an \(r^{-3.5}\) term;
\item
  \textbf{L3, hidden-charge existence.} Some configurations cannot be fit with the corrected force alone; there must be an unobserved property on each body;
\item
  \textbf{L4, hidden force law.} The new force depends on the product of two charges and a periodic distance term;
\end{itemize}
and report consecutive depth: credit up to the first miss along L1 to L4. ``Highest layer passed'' misleads, because layers can fire non-consecutively for the wrong reasons; any charge-product term earns automatic credit at the parity-symmetry layer regardless of the chain beneath it. We corrected this definition before the measured runs.

The target is therefore structural discovery, not direct numerical recovery of \(s_1\) and \(s_2\).  These symbols denote an unobserved scalar property attached to each body.  L1 and L2 concern the visible radial law; L3 asks whether the agent infers that a latent per-body property is needed; and L4 asks whether it recovers the product coupling and periodic distance dependence.  Since the dynamics contain \(s_1\) and \(s_2\) only through their product, the two individual values are never separately identifiable from trajectories alone.

Two further layers exist: L5, the conserved quantity \(Q\), and L6, the discrete parity-plus-charge-sign symmetry. Both are orthogonal to a force-law prompt and not reliably elicited by it. We score and report them, but the capability measure throughout is consecutive L1 to L4 depth.

\subsection{The observability switch}\label{the-observability-switch}

We can turn recoverability of L3 and L4 on and off deterministically. The construction is portable to any parametric law.

\begin{enumerate}
\def\labelenumi{\arabic{enumi}.}
\tightlist
\item
  Pick a parameter that enters the law only through a symmetric combination with another parameter (a product, a sum, a ratio).
\item
  For the degenerate arm, observe a single configuration. The combination is then one constant, and the individual factors have no separate effect on any observable.
\item
  For the observable arm, observe a family of systems with the same law and couplings but distinct individual factors. Across the family the combination varies, leaving a signature no single global constant can absorb.
\item
  Before any agent runs, verify the switch with the diagnostic of Section~\ref{the-diagnostic}: rank drop and named null direction in the degenerate arm, full rank in the observable arm.
\end{enumerate}

Two things change between arms in our runs, not one: the set of systems shown to the agents, and the task text that introduces them. The family arm's prompt states that the systems differ in one unobserved per-body property and asks the agents to denote it $s_1, s_2$; the single-configuration prompt does not (Appendix~\ref{appendix-prompts} reproduces both). A clean switch holds the task text fixed; ours did not, and Section~\ref{observability-lift-and-second-regime-transfer} reads the result accordingly.

In the Alien Universe the combination is \(s_1 s_2\). At the tuned operating point both bodies carry equal unit charges, so \(s_1 s_2 \equiv 1\) and \(\kappa s_1 s_2\) collapse to constants. A deterministic, no-LLM audit established three facts. The with-charge and without-charge acceleration-fit columns are byte-identical. The fitted oscillatory amplitude tracks only the charge product, so (1, 1) and (2, 0.5) are observationally indistinguishable. Equal-product configurations integrate to bit-identical trajectories. The data constrain the lumped quantity \(\beta s_1 s_2\) and nothing finer. The oscillatory force and \(\gamma\) are identifiable as phenomena, but attributing them to a per-body charge is not singled out by the data. Thus, the single-configuration protocol makes the L3--L4 hidden-property interpretation non-identifiable, while the family protocol supplies the cross-system variation needed to test it.  Even the family identifies only the latent product-coupling structure, not the two charges separately. L3 and L4 are verifiable but not discoverable at equal charges; depth 2 is the discoverable ceiling there.

The observable arm hands the agents a family with identical \(\alpha\), \(\beta\), \(\gamma\), \(\kappa\) and distinct per-body charges, so amplitude and tangential coupling vary with the per-system product. The L3 and L4 gates credit the hidden charge only when per-configuration charge-product scaling does material residual work beyond one global amplitude, so a single global \(\beta\) cannot pass L3 by accident.

The diagnostic of Section~\ref{the-diagnostic} returns the \((1, -1)\) null direction over \((s_1, s_2)\). The swap \(s_1 \leftrightarrow s_2\) is a symmetry of every design, including a family, because only the product enters. The family does not separate \(\beta\) from the charges either: with every system's charges unknown, rescaling all charges by \(\lambda\) and \(\beta\) by \(\lambda^{-2}\) leaves every system unchanged, so the family identifies the per-system products \(\beta s_1 s_2\), their ratios across systems, and nothing finer (rank 2 of 5 for a two-system family over \((\beta, s_1, s_2, t_1, t_2)\)). The certificate of Section~\ref{certified-exhaustion-and-plateau-triage} formalizes this; the correct target is the product, not the factors.

\subsection{What the agents see}\label{what-the-agents-see}

Observation is data-only, whether passive or requested. Every observed system arrives as a bundle of times, positions, velocities, and masses from the simulator. The charges, the couplings \(\alpha\), \(\beta\), \(\gamma\), \(\kappa\), the conserved quantity \(Q\), the parity symmetry, and the closed-form law are never returned. A requested measurement, whether a new initial condition or a family member, returns the same bundle. The simulator stays the oracle and the answer key stays hidden.

Scoring runs separately with no LLM, against held-out initial conditions the agents never see: symbolic equivalence via sympy, numerical agreement of forward-integrated trajectories on the held-out set, and a qualitative shape check. The agent cannot game data it has never fit.

\subsection{Construct-validity note}\label{construct-validity-note}

Two scope conditions apply to everything measured on the Alien Universe. First, the Alien Universe is a synthetic model we authored. Obviously, reaching depth on it is a capability measurement under controlled observability, not new knowledge about nature. The invented law makes the switch exact and the answer key machine-checkable with no LLM in the loop.

Second, the single-configuration L1 and L2 gates are narrower than they appear. Their numerical halves are computed from the recovered truth acceleration on a fixed basis, without reference to the proposal's own terms. Each reduces to a structural test: an \(r\)-exponent within \(\pm 0.3\) of the true 3.5, with sign and coefficient not gated. We confirmed this with a post hoc rescore of stored proposals under a proposal-coupled L2 gate that least-squares-fits each proposal's own correction exponent to the held-out acceleration. Under that gate the architecture separation reported in Section~\ref{architecture-versus-tools-a-gate-dependent-separation} collapses: both arms pass L2. Over the orbit range \(r \in [1, 1.41]\) the solo agent's \(r^{-3}\) term reduces the post-Newton residual marginally more than the Council's \(r^{-3.4}\) (0.184 versus 0.172). The data-optimal correction exponent is about 2.95 over the orbit range and about 3.40 over clean small-radius drops. Over those drops the residual reduction saturates near 99\% for every exponent, so any threshold gate is vacuous there. With a free-symbol coefficient, a per-proposal numeric gate must fit it against held-out data, and the fit absorbs the sign; even the true \(r^{-3.5}\) correction fits a negative coefficient. Sign and coefficient therefore cannot be gated at the single-configuration level.
The cross-configuration L3 gate is different in kind. It credits the hidden charge only when a proposal recovers the coupling from a family with distinct charges, work a single degenerate configuration cannot fake. That is where the observability claim of Section~\ref{observability-lift-and-second-regime-transfer} is located. Every gate, including the rescore, is deterministic code; no model judges ground truth. This bounds how the depth number may be read, and Section~\ref{the-diagnostic-turned-on-the-benchmarks-own-metric} turns the diagnostic on the L2 metric's observation design.


\section{Results}\label{results}

Contribution 1, the verdict and certified triage, is tested on systems we did not build (Section~\ref{the-diagnostic-on-systems-we-did-not-build}) and in a live loop (Section~\ref{the-loop-closes-and-closes-by-proof}). Contribution 2, the Alien Universe switch, is tested in Section~\ref{observability-lift-and-second-regime-transfer}; contribution 3, the benchmark's own metric, in Section~\ref{the-diagnostic-turned-on-the-benchmarks-own-metric}; and contribution 4, validation beyond our testbed, in Sections~\ref{the-diagnostic-on-systems-we-did-not-build}, \ref{transfer-to-integer-sequences-with-a-machine-checked-certificate}, \ref{the-diagnostic-on-odebench}, \ref{criterion-head-to-head}, and~\ref{discoverphysics-audit}. The architecture comparison of Section~\ref{architecture-versus-tools-a-gate-dependent-separation} is secondary; its detail is in Appendix~\ref{appendixtoka-architecture-versus-tools-in-detail}.

\subsection{Protocol}\label{protocol}

Every measured cell follows one protocol, fixed before any run. A cell is eight seeds, and the two arms of an ablation are paired on the seed. We report the mean paired difference with a 95\% bootstrap confidence interval, a one-sided Wilcoxon signed-rank test in the pre-registered direction, and Cliff's delta, with a Holm-Bonferroni correction within each prediction family. At $n = 8$ the one-sided Wilcoxon test bottoms out at $p = 1/2^8 = 0.0039$ when all pairs favor one arm and at $p = 1/2^7 = 0.0078$ with one tie. Several results sit at these floors, so we read effect sizes alongside p-values. In one case (Section~\ref{architecture-versus-tools-a-gate-dependent-separation}) the registered correction plan cannot reach significance at all.

Three rails apply to every run.

\textbf{Pre-registration.} For every live-agent measurement and for the ODEBench, criterion-comparison, and DiscoverPhysics analyses, each prediction, with its quantitative bands and falsification condition, was committed to the repository before the run; the ordering is verifiable in the development repository's history, available on request. Two amendments were logged after some data were in hand and before the runs they govern (a roster change for EXP-089 and a numerics change for EXP-261). The depth-ladder definition was corrected before the measured batch but after three treatment-arm pilot runs. The textbook ODE table, the pharmacokinetic case, the constructed battery, and the integer-sequence runs are exploratory. Failed predictions are reported as failed.

\textbf{Deterministic scoring with no language model.} A sympy and scipy procedure walks the consecutive ladder of Section~\ref{the-depth-ladder} and stops at the first layer a run's hypothesis set fails to clear. A grep gate fails the pipeline if a model call appears in the scoring path.

\textbf{Exploration and measurement are separated.} Agents see only times, positions, velocities, and masses, never the force law. Depth is scored on held-out initial conditions the agent cannot select, and the cross-configuration charge gate on held-out charge products strictly off the observed family.

Two recorded failures preceded the measured results. In the first (EXP-072), the Council ran on passive orbit data in two runs, reliably rediscovered the inverse-square force, and reached only depth 1, falsifying the pre-registered depth-2 prediction. Passive trajectory data does not surface the correction term, which is why the active-experimentation battery exists. In the second (EXP-074), the eight-seed test of that battery was confounded by infrastructure on both attempts: provider quota exhaustion, then a fallback provider that could not emit valid hypothesis JSON. We record it as a confounded null; the mechanism was validated in isolation and the powered measurement moved to the roster of Section~\ref{observability-lift-and-second-regime-transfer}.

Provider rosters changed between experiments. The observability lift (Section~\ref{observability-lift-and-second-regime-transfer}) ran on a mixed Gemini and NVIDIA Council, the same in both arms. The architecture ablation, the live loop, and the transfer runs ran all-Gemini, with the same tool stack on a different roster.

\subsection{The diagnostic on systems we did not build}\label{the-diagnostic-on-systems-we-did-not-build}

Every Alien Universe result rests on physics we invented, and an engine can be tuned to its authors' testbed without anyone intending it. We therefore pointed the same unmodified code, with no language model, at systems with settled identifiability answers, using the sensitivity-matrix primitive (local) and the exact symbolic checker (global).

\textbf{Classic ODEs.} We run Lotka-Volterra, Van der Pol, and Lorenz, each under a full-state and a single-coordinate design, with initial conditions known (Table~\ref{tab:ode}). For each cell we integrate with scipy LSODA (rtol = atol = 1e-12) on a fixed time grid, central-difference the flattened observations at the nominal point, and read off rank, verdict, and condition number.

\small\begin{longtable}[]{@{}
  >{\raggedright\arraybackslash}p{(\columnwidth - 10\tabcolsep) * \real{0.22}}
  >{\raggedright\arraybackslash}p{(\columnwidth - 10\tabcolsep) * \real{0.18}}
  >{\raggedleft\arraybackslash}p{(\columnwidth - 10\tabcolsep) * \real{0.12}}
  >{\raggedleft\arraybackslash}p{(\columnwidth - 10\tabcolsep) * \real{0.09}}
  >{\centering\arraybackslash}p{(\columnwidth - 10\tabcolsep) * \real{0.23}}
  >{\raggedleft\arraybackslash}p{(\columnwidth - 10\tabcolsep) * \real{0.16}}@{}}
\caption{The diagnostic on textbook ODE systems with initial conditions known: rank and condition number of the sensitivity matrix at the nominal parameters.}\label{tab:ode}\\
\toprule\noalign{}
\begin{minipage}[b]{\linewidth}\raggedright
System
\end{minipage} & \begin{minipage}[b]{\linewidth}\raggedright
Design
\end{minipage} & \begin{minipage}[b]{\linewidth}\raggedleft
n\_params
\end{minipage} & \begin{minipage}[b]{\linewidth}\raggedleft
rank
\end{minipage} & \begin{minipage}[b]{\linewidth}\centering
identifiable? (initial conditions known)
\end{minipage} & \begin{minipage}[b]{\linewidth}\raggedleft
cond. number
\end{minipage} \\
\midrule\noalign{}
\endhead
\bottomrule\noalign{}
\endlastfoot
Lotka-Volterra (4-param) & full-state (x, y) & 4 & 4 & yes & 21 \\
Lotka-Volterra (4-param) & prey-only (x) & 4 & 4 & yes & 169 \\
Van der Pol & full-state (x, y) & 1 & 1 & yes & 1 \\
Van der Pol & position-only (x) & 1 & 1 & yes & 1 \\
Lorenz & full-state (x, y, z) & 3 & 3 & yes & 23 \\
Lorenz & x-only (x) & 3 & 3 & yes & 25 \\
\end{longtable}\normalsize

Lotka-Volterra uses $\bth = (1, 1, 1, 1)$, initial condition $(1, 0.5)$, and $t \in [0, 20]$; Van der Pol uses $\mu = 0.5$, $(2, 0)$, and $t \in [0, 20]$; Lorenz uses $(\sigma, \beta, \rho) = (10, 8/3, 28)$, $(-8, 8, 27)$, and $t \in [0, 3]$, a pre-divergence window, since a local linearization of a chaotic system is meaningful only within a few Lyapunov times. We use the four-parameter Lotka-Volterra $(\alpha, \beta, \delta, \gamma)$ rather than the two-parameter pysindy form in our registry, which ties the rates together. The exact symbolic checker agrees on the full-state vector fields: all are structurally identifiable with no confounded groups.

The prey-only Lotka-Volterra row depends on what is assumed known. With the initial condition known, the four rates are locally identifiable from prey alone (rank 4 of 4), at a condition number above the full-state value that grows with the window (124, 169, and 361 for horizons $[0, 4]$, $[0, 20]$, and $[0, 40]$). With the predator initial condition $x_2(0)$ unknown, StructuralIdentifiability.jl reports $b$ and $x_2$ non-identifiable through the scaling $x_2 \to \lambda x_2$, $b \to b/\lambda$, with only $b\,x_2$ identifiable \cite{structidjl}. Our check holds the initial condition fixed and does not see this (Section~\ref{limitations}).

\textbf{A planted degeneracy.} In a Lotka-Volterra variant where $\alpha$ and $\beta$ enter only through their product and $\delta$ and $\gamma$ only through their sum, the diagnostic returned rank 2 of 4 with confounded directions $(-0.707, +0.707)$ on $(\alpha, \beta)$ and $(+0.707, -0.707)$ on $(\delta, \gamma)$, the exact signatures of the planted confounds.

\textbf{A published non-identifiability (exploratory).} The oral one-compartment pharmacokinetic model is the textbook case of parameters that enter only as a ratio \cite{rowland-tozer-2011, gabrielsson-weiner-2016}: $y(t) = \frac{F D k_a}{V (k_a - k_e)} \left(e^{-k_e t} - e^{-k_a t}\right)$, with $\bth = (k_a, k_e, V, F)$ and dose $D$ known. With 24 samples over 24 h at $\bth = (1.2, 0.25, 30, 0.6)$, the diagnostic returns rank 3 of 4 with the confounded direction along the $(V, F)$ scaling axis (cosine $1.000000$), and the symbolic checker reports the confounded group $\{F, V\}$. Of three follow-up designs, four-times-denser oral sampling and a 72 h oral window stay at rank 3 of 4, while adding an intravenous arm, where $F = 1$ by definition, reaches rank 4 of 4 (condition number $1.2\times10^{2}$). The ranker recommends the intravenous arm, the absolute-bioavailability design pharmacologists use \cite{rowland-tozer-2011}. The model also has the flip-flop symmetry: swapping $k_a$ and $k_e$ with $F$ rescaled by $k_a/k_e$ gives identical oral observations (maximum difference 0.0) and the same local verdict at both points, so the local check cannot see the second solution. The intravenous arm pins $k_e$ and breaks the swap (maximum difference 1.7).

\textbf{Where the symbolic check ends.} The symbolic checker originally resolved confounds pairwise, asking whether two parameters' sensitivity functions are proportional. On a battery of eight model classes with known answers (EXP-082) it was exact on seven, including $A \sin(\omega r)$ (identifiable) and $A k e^{-r}$ (confounded). It missed a higher-order confound: in $f = a r + b r^2 + c (r + r^2)$ only $a + c$ and $b + c$ are recoverable, yet no pair of sensitivities is proportional, while the sensitivity rank returns 2 of 3 with null direction $(1, 1, -1)$. After the ODEBench sweep met the same blind spot (Section~\ref{the-diagnostic-on-odebench}), the checker gained an exact test for constant-coefficient linear dependence among the derivative functions, evaluated at fifty digits at two independent generic rational parameter draws; it finds both cases. The symbolic method never falsely claims non-identifiability, and the numerical path remains the complement for black-box hypotheses and for practical (sloppy) degeneracy. A constructed battery characterizes the method, not a distribution over real problems.

\subsection{Observability lift and second-regime transfer}\label{observability-lift-and-second-regime-transfer}

\begin{figure}[t]\centering
\includegraphics[width=0.62\linewidth]{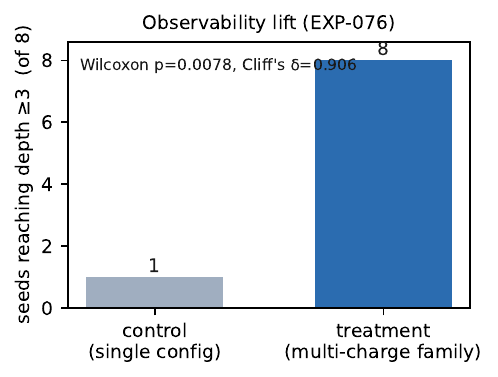}
\caption{Observability lift on the Alien Universe (EXP-076): the made-observable treatment reaches discovery depth $\geq 3$ on 8/8 seeds against 1/8 for the single-configuration control. The treatment also carries the fitting engine; the factorial (EXP-089) attributes the cross-configuration layer to the family and the correction-exponent layer to the engine.}\label{fig:obs}
\end{figure}

\begin{figure}[t]\centering
\includegraphics[width=0.62\linewidth]{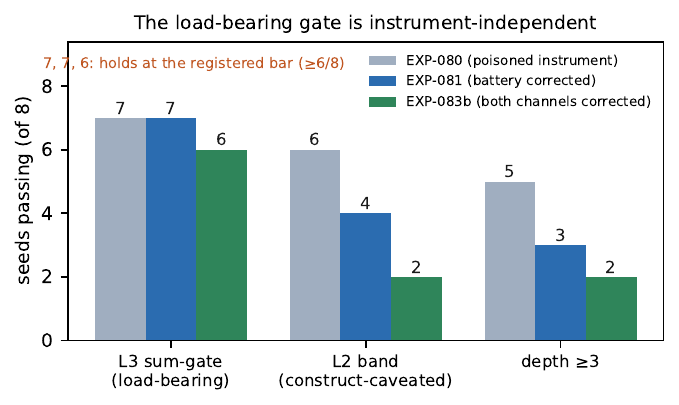}
\caption{The cross-configuration sum gate (L3) on the second Alien Universe regime holds at 7/8 under a corrupted in-loop fitting engine (EXP-080), 7/8 with the probe battery corrected (EXP-081), and 6/8, its pre-registered bar, with both agent-visible channels corrected under clean weather (EXP-083b). The band-gated layers fall across the same runs. A fourth, outage-confounded run (EXP-083) is reported in the text and excluded here.}\label{fig:transfer}
\end{figure}

\textbf{The lift (EXP-076, Fig.~\ref{fig:obs}).} This eight-seed paired ablation is scored post hoc by consecutive depth (Section~\ref{the-depth-ladder}) on each run's full hypothesis set. The treatment arm sees a family of two-body systems sharing one law with distinct per-body charges, plus residual feedback and the exponent-fitting battery; the control sees a single charge configuration, where the charge product is degenerate, with both tools disabled. The treatment reaches depth at least three on 8/8 seeds (six at depth three, two at depth four; mean 3.25), the control on 1/8 (mean 1.875): one-sided Wilcoxon $p = 0.0078$ with one tied pair, Cliff's delta 0.906, and mean paired difference 1.375 (bootstrap 95\% CI [0.75, 2.00]). The one control seed at depth three was credited by the looser single-configuration L3 gate; under a symmetric control-L3 gate applied post hoc it drops out (0/8, delta 1.0, $p = 0.0039$). We report both.

Three facts bound this comparison. The single-configuration arm cannot pass the cross-configuration L3 gate, which needs two distinct charge products. The family arm's task text states that a per-body property exists and names it (Section~\ref{the-observability-switch}), which is most of what L3 asks; what remains is to find a product coupling that does residual work on held-out charge products. And runs share distilled judging criteria (Section~\ref{the-loop}), so seeds are not fully independent. The lift therefore confirms that the switch behaves as designed: when the coupling is identifiable and the task points at it, the Council recovers it on every seed, and otherwise it cannot. It is not evidence of unprompted discovery, and protocol and prompt effects are not separated.

The treatment arm also changed three factors at once: the family, residual feedback, and the fitting battery. We ran the two missing cells of the factorial (EXP-089), eight seeds each, pre-registered with four falsifiers: family-only (feedback and battery off) and engine-only (single configuration, feedback and battery on). Family-only passes L3 on 8/8 seeds and L2 on 2/8; engine-only passes L2 on 8/8 and L3 on 1/8. Across the four cells, L3 credit is 16/16 with the family against 4/16 without (Fisher two-sided $p = 1.6\times10^{-5}$; the single-configuration credits come from the looser gate, so this also compares two gates), and L2 credit is 16/16 with the fitting engine against 8/16 without ($p = 0.0024$). Consecutive depth three is reached by 2/8 family-only and 1/8 engine-only seeds, against 8/8 with both. All four predictions held. The family, with its task text, makes the L3 gate passable, as it must by construction; the fitting engine delivers the exponent and nothing else. Seeds 0 to 3 ran on the registered mixed roster and seeds 4 to 7 all-Gemini, after the NVIDIA model reached end of life; the roster was identical across arms and the change was logged as an amendment. The contrasts hold within each stratum, but the strata differ: family-only passed L2 on 0/4 mixed-roster and 2/4 all-Gemini seeds, engine-only passed L3 on 1/4 and 0/4, and both family-only depth-three seeds are all-Gemini.

\textbf{Second-regime transfer (Fig.~\ref{fig:transfer}).} We re-ran the unchanged stack on a second regime with a sum coupling $s_1 + s_2$ and correction term $r^{-2.5}$, outside the bands both tools were tuned on; transfer and non-transfer were both committed as publishable before launch. In the first run (EXP-080), depth at least three fired on 5/8 seeds, exactly the registered bar (exact-binomial 95\% CI [0.24, 0.92]), L2 on 6/8, and the L3 sum gate on 7/8. The probe battery, however, simulated the first regime and told the agents the correction fit best near 3.4. With the battery corrected (EXP-081), L3 held at 7/8, L2 fell to 4/8, and depth at least three fell to 3/8, below the bar, with heavily overlapping intervals, while mean terminal exponent error under the regime-aware rescore improved from 0.56 to 0.40. The corrected battery reports the data-optimal exponent, about 2.1 (biased below 2.5 by the hidden charge term), and the L2 band is centered on the truth at [2.2, 2.8], so a more data-faithful engine scores worse on a truth-centered band. With the second agent-visible channel, the residual-feedback held-out set, also corrected, EXP-083 missed its registered L3 bar (4/8) during a roughly six-hour provider outage that starved four seeds. The re-run with the same bar (EXP-083b, registered minutes after EXP-083's outcome, and to be weighed as a retry after a failure) met it at 6/8 (95\% CI [0.35, 0.97]), with depth at least three and L2 each at 2/8; its two L3 misses coincide with two seeds hit by a wall-clock stall.

The band-gated layers fell 5/8, 3/8, and 2/8 as the engine was corrected, a negative result about band metrics under hidden-term contamination rather than a transfer of the correction layer. What transferred is the proposal-coupled signal: the L3 gate held across every engine state tested, and the structural diagnostic confounds $\{s_1, s_2\}$ for both the product and the sum coupling with no new code.

\subsection{The loop closes, and can stop with a certificate}\label{the-loop-closes-and-closes-by-proof}

\begin{figure}[t]\centering
\includegraphics[width=0.98\linewidth]{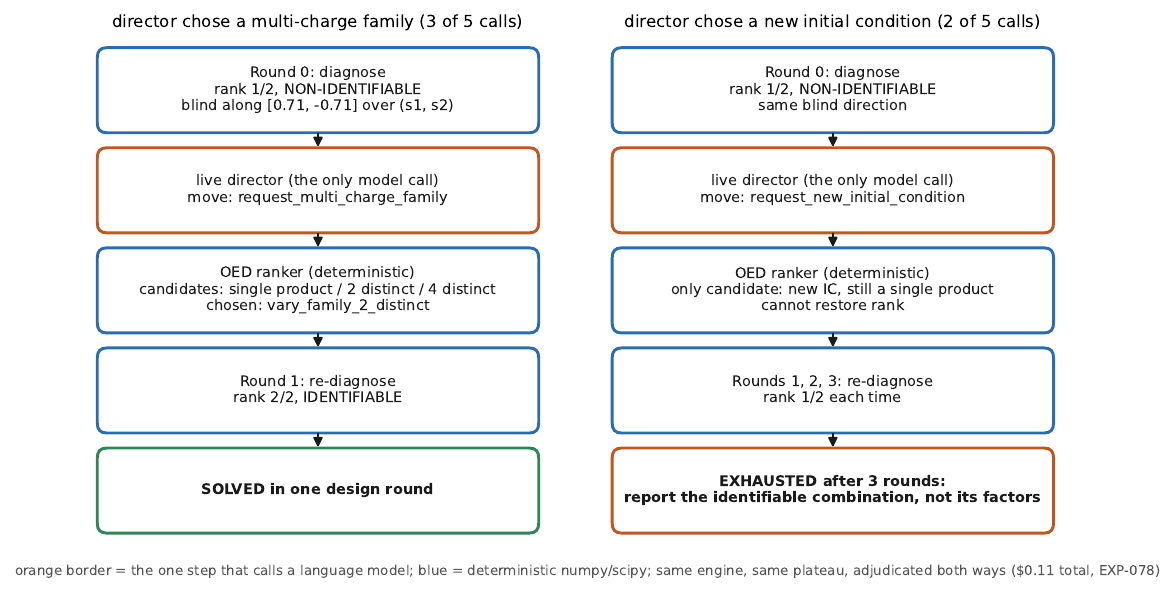}
\caption{The identifiability loop closed by a live agent decision, both ways (EXP-078). Left: the director requests a multi-charge family, the deterministic ranker selects the two-distinct-product design, and the re-diagnosis returns full rank (SOLVED). Right: the director requests a new initial condition, no candidate restores rank, and the loop reports EXHAUSTED. With the certificate of Section~\ref{certified-exhaustion-and-plateau-triage} the right-hand branch terminates at round 0 by proof.}\label{fig:loop}
\end{figure}

We ran the loop of Algorithm~\ref{alg:identifiability-loop} once with a live proposer (EXP-078, Fig.~\ref{fig:loop}) on the amplitude model $A(q) = \beta q^{p}$ over charge products $q = s_1 s_2$, with $\bth = (\beta, p)$. One product gives rank 1 of 2 and two distinct products give rank 2 of 2. This model is separate from the force-law diagnostic of Section~\ref{the-diagnostic}, whose null direction lies along the charge axis.

The synthesizer, acting as director, saw a constructed plateau (one distinct form, dominant share one, flat residual) and chose between two fixed moves. On 3 of 5 scored calls it requested a multi-charge family; the ranker chose a two-product design, re-diagnosis returned rank 2 of 2, and the loop returned SOLVED. On the other 2 it requested a new initial condition; no single-product design restored rank, and the loop returned EXHAUSTED after 3 rounds. An earlier probe call also chose a new initial condition (six live calls, three family moves). Only the director called a language model, and the leak audit held. Its prompt carried categorical plateau facts and the name of the unidentifiable direction, but no charge, coupling, exponent, or law value, and its rationales (``a family of varied systems breaks the degeneracy''; ``a new initial condition will generate a different trajectory'') never named the hidden charge.

The certificate turns EXHAUSTED from a budget verdict into a proof at the operating point (Proposition~1). On the full parameterization $\bth = (\beta, s_1, s_2)$ of Section~\ref{the-law-as-simulated} and the same single-configuration design it finds two null directions. The scaling direction $(\beta, -s_1/2, -s_2/2)$ is a symmetry of every single-configuration design, so with the certificate in the loop the new-initial-condition branch closes at round 0 with a symbolic certificate at a verified regular point (\texttt{exhausted\_by\_certificate}). The swap direction $(0, 1, -1)$ is a symmetry of every design, families included, since the law sees only the charge product. A family breaks the scaling only if some charge is known (Section~\ref{the-observability-switch}); what it breaks is the amplitude model's confound, which is why the family branch returns \texttt{solved} on that model.

A pre-registered held-out test of the numeric route on Michaelis-Menten kinetics (EXP-087) gave rank 2 of 3 from substrate time courses, with the null direction along the $(k_{\mathrm{cat}}, E_0)$ scaling axis. The certificate held over 32 sampled designs at the registered bar (normalized stacked response at most $10^{-7}$), the triage returned \texttt{change\_class}, and an enzyme-concentration assay restored full rank. On a second platform (Linux x86-64, numpy 2.4.4, scipy 1.17.1) the stacked response is $1.2 \times 10^{-7}$, just above the bar, although the next singular value is $3.3 \times 10^{-3}$. The bar sits at the integrator's finite-difference floor (Section~\ref{the-diagnostic-on-odebench}), so we count the prediction as platform-dependent.

The same hook fires on the pharmacokinetic model of Section~\ref{the-diagnostic-on-systems-we-did-not-build}: restricted to oral sampling schedules, it certifies the $(V, F)$ direction for the whole class and stops at round 0 with \texttt{change\_class}; only the intravenous arm, another class, breaks it. The narrow-window sequence ansatz of Section~\ref{transfer-to-integer-sequences-with-a-machine-checked-certificate} collapses through the window, not the class, so it is not certified and the ranker resolves it in one round. Neither MDA nor LLM-ACES, as published, returns such a verdict. With $n = 5$, a fixed two-move vocabulary, and a constructed plateau, this is a capability demonstration and the choice split is descriptive.

\subsection{The diagnostic turned on the benchmark's own metric}\label{the-diagnostic-turned-on-the-benchmarks-own-metric}

\begin{figure}[t]\centering
\includegraphics[width=0.62\linewidth]{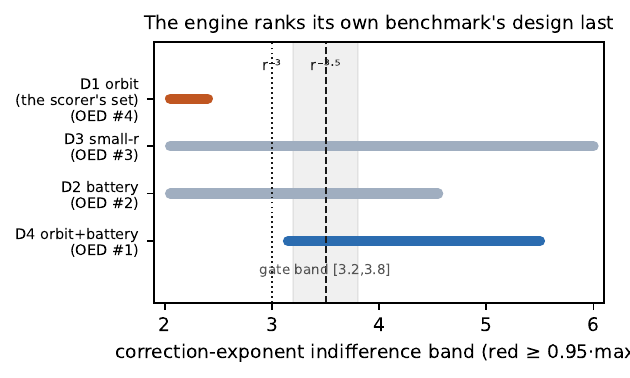}
\caption{The engine diagnoses its own benchmark. With the correction exponent as a free parameter, the design ranker places the depth scorer's own observation design last of four; its indifference band (orange) does not reach the gate band, while the top-ranked design (blue) excludes $r^{-3}$.}\label{fig:oed}
\end{figure}

The depth metric cannot distinguish a correction exponent of 3.4 from the true 3.5 on its own held-out data (Section~\ref{construct-validity-note}). We made this formal with the diagnostic, with no LLM call and predictions committed beforehand in a dated memo (EXP-088).

\textbf{Setup.} We treat the exponent as a free parameter, $a(r; c_0, c_1, e) = -c_0/r^2 - c_1/r^{e}$ with $\bth = (c_0, c_1, e)$, and run the unchanged diagnostic and ranker over four designs: D1, the held-out orbit set the depth scorer grades on; D2, the agents' clean radial-drop battery; D3, its small-radius window; and D4, the union of D1 and D2. For each we compute an indifference band: the exponents reaching at least 95\% of the best attainable residual reduction.

\textbf{Three predictions failed.} We predicted that D1 would be markedly worse conditioned than the battery designs, with a sloppy $(c_1, e)$ direction. Instead every design is rank 3 of 3, with condition numbers between $1.5\times10^{2}$ and $5.7\times10^{2}$; the collapse is not a local effect, consistent with the local-versus-global boundary of Section~\ref{the-diagnostic-on-systems-we-did-not-build}. We predicted that D1's band would be too wide; it is displaced instead. Over the scorer's full held-out span, $r \in [0.90, 2.32]$ (wider than the orbit range quoted in Section~\ref{construct-validity-note}), D1's band is $e \in [2.05, 2.40]$, disjoint from the gate's band, so the gate passes a 3.4 proposal because its 5\% threshold is permissive, not because the data prefer 3.5. We predicted that no design's band would exclude $r^{-3}$; D4's band, [3.15, 5.50], does, while keeping $r^{-3.5}$.

\textbf{One prediction held.} The ranker places D1 last. All four designs have full rank, so the condition-number tie-break decides, with thin margins (D4 and D2 differ in the fourth significant figure of the score), and the ranker, unprompted, puts D4 first. Under the registered falsification clause, D4 is the design for any tightened-gate re-measurement of Section~\ref{architecture-versus-tools-a-gate-dependent-separation}. Figure~\ref{fig:oed} plots all bands against the gate's. D4 rejects $r^{-3}$ but cannot pin 3.5 against steeper exponents, since its band runs to 5.50. The general lesson is to write the quantity a metric should resolve as a free parameter, check its identifiability on the metric's own observation design, and compute the indifference band before paying for a measurement.

\subsection{Architecture versus tools: a gate-dependent separation}\label{architecture-versus-tools-a-gate-dependent-separation}

\begin{figure}[t]\centering
\includegraphics[width=0.62\linewidth]{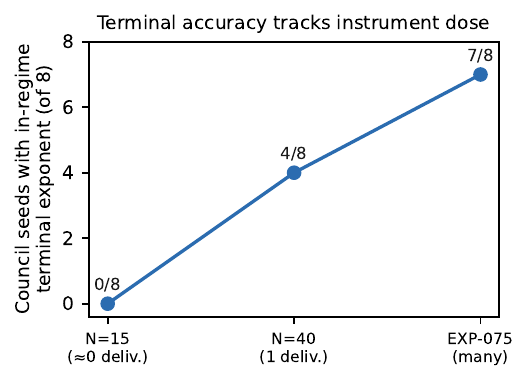}
\caption{Terminal structural accuracy tracks the engine-feedback dose rather than the architecture as such: near-zero deliveries (N=15: two across the eight Council seeds) yield zero in-regime terminals, one delivery (N=40) about half, many (EXP-075) nearly all.}\label{fig:dose}
\end{figure}

The five-agent Council and a single generalist agent (Solo) ran on the same all-Gemini tool stack, eight seeds each (EXP-075). Under the registered band gate the Council reached depth 3 and the Solo depth 1 on every seed, a maximal separation at the $n = 8$ floor (Appendix~\ref{a.1-the-equal-wall-clock-comparison}), carried entirely by the L2 layer. Under the proposal-coupled gate of Section~\ref{construct-validity-note} the separation vanishes (Appendix~\ref{a.3-the-proposal-coupled-rescore}).

What remains is a modest structural residual: the Council's correction exponents cluster near 3.4 and the Solo's near 3.0. The gate-free terminal-exponent error (the distance of the final persisted hypothesis's exponent from the truth) separates the arms on the exploration set (mean 0.15 versus 0.50, one-sided $p = 0.0078$). The pre-registered confirmatory test on the call-matched $N = 40$ cell misses ($p = 0.0625$; four pairs favor the Council, none the Solo, four tie), so by our registered falsifier the claim is not a result. The median error over every distinct exponent an arm proposed shows no separation in either dataset: both arms propose broadly, and what differs is where the loop's final state lands.

The metric did expose a mechanism (Fig.~\ref{fig:dose}): terminal accuracy tracks how often the end-of-round step delivered the fitting engine's feedback. With near-zero deliveries at $N = 15$ (two across the eight Council seeds, none for the Solo), 0 of 8 terminals were in regime; with one delivery on seven of eight Council seeds at $N = 40$ (one in total for the Solo), 4 of 8; with many deliveries in the wall-clock cell, 7 of 8. The call-matched control (EXP-077) caps both arms at the same number of calls. At $N = 15$ the arms tie at depth 1; at $N = 40$ the gap partly returns for the Council while the Solo stays at depth 1, and the registered Holm plan is unsatisfiable (Appendix~\ref{a.2-the-call-matched-control}, with the partial $N = 130$ cell). We read the comparison descriptively: the deterministic engine already knows the exponent, and what the architecture measurably contributes here is keeping its feedback in the loop.

\subsection{Transfer to integer sequences with a machine-checked certificate}\label{transfer-to-integer-sequences-with-a-machine-checked-certificate}

The last test asks whether the diagnostic depends on anything physical. A small adapter, with nothing inside the engine of Section~\ref{the-identifiability-engine} changed, replaces the simulator with an exact integer-sequence generator and the verifier with a sympy check that a candidate closed form minus the sequence reduces to zero for all \(n\). The predict closure works in log space, so an ansatz \(f(n;\theta)\) evaluated at sample points is a map from parameters to observations. The diagnostic, ranker, and loop run verbatim, and a test pins that the diagnostics package imports nothing from the sequence package.

In the blind case (EXP-084) the ansatz \(f(n;\theta) = \theta_0 \binom{2n}{n} n^{\theta_1} \theta_2^n\) targets \(\sum_k \binom{n}{k}^2\), with the closed form stored nowhere in the loop's inputs. From a three-point tail window the diagnostic reported rank 2 of 3: over a short window at large \(n\), \(\log n\) is numerically linear in \(n\) and the sensitivity columns collapse, the same rank drop the charge product causes in the Alien Universe, here from a poor sample design. The ranker chose a geometric spread over a denser tail (condition number $2.9\times10^{2}$ against $3.9\times10^{6}$); the refit reached rank 3 of 3 and recovered \(\theta = (1, 0, 1)\) exactly, and sympy certified the Vandermonde identity \(\sum_k \binom{n}{k}^2 = \binom{2n}{n}\) for all \(n\). The same ansatz on the central Delannoy numbers, a control, produced no exact snap and was not certified.

We then extended the ansatz space to linear combinations over a graded basis of combinatorial atoms and targeted three sums (EXP-085): two moments of the discrete arcsine law and a shifted convolution from a known tail-sum family (Appendix~\ref{appendixtokb-closed-form-derivations-and-search-trail}). The engine derived each form by an exact rational solve from five, six, and seven small samples. Certification uses a generating-function certificate: a weight \(k^j\) becomes the operator \((x\,d/dx)^j\), a convolution becomes a product of algebraic factors, and sympy decides whether the difference from the assembled candidate reduces to literal zero. Across every family tested the pipeline issued no false certificate. One true identity, \(\sum_k k\binom{n}{k} = n2^{n-1}\), came back UNVERIFIED on a sympy non-decision (Appendix~\ref{appendixtokb-closed-form-derivations-and-search-trail}), a false negative. On the pre-committed control, a mixed-singularity family outside the basis, the exact solve found no rational solution and refused; that is out-of-basis detection and says nothing about whether a closed form exists. The certifier is sound when it reaches zero and incomplete otherwise.

\subsection{The diagnostic on the benchmark a competitor uses}\label{the-diagnostic-on-odebench}

Eight seeds on a testbed we built is little volume next to the 122 ODE systems of LLM-ACES. Because the diagnostic is deterministic and free, we ran it over the LLM-ACES benchmark itself, on the exact data LLM-ACES hands its proposer, with predictions committed first (EXP-261). The task set is the 63 ODEBench systems in the pinned LLM-ACES release (commit 60d4df7); the other 59 (ODEBase) are not in that release. The round-0 data are one noise-free trajectory from the published first initial condition on $t \in [0, 1]$ at 100 uniform points, with every state and its analytic derivative observed. The acquisition class is what LLM-ACES can do next: one added trajectory from an initial condition in its per-system box (ten candidates) plus the benchmark's held-out second initial condition. For each system we compute the exact symbolic structural verdict, the local rank and condition number at the published constants from forward sensitivities, a symbolic certificate for each null direction (is the vector field invariant along it for every state?), and the ranker's best added design. The run takes minutes with no language model; the per-system table is a supplementary file.

\textbf{Result.} One system has no constants. Of the other 62, 60 are identifiable at round 0 (exact binomial 95\% CI 0.89 to 1.00; median condition number 12), none is practically non-identifiable, and two are rank-deficient in raw coordinates. The RC circuit \(\dot{x} = (c_0 - x/c_1)/c_2\) has three constants that enter only as \(c_0/c_2\) and \(1/(c_1 c_2)\). The symbolic checker finds one exact dependence among the derivative functions, the sensitivity rank is 2 of 3, and the certificate proves the vector field invariant along the null direction for every state, so no initial condition and no time window can separate the constants: the triage is change the class. The best added design leaves rank 2 of 3, and two models displaced along the null direction give rollouts that agree to $5\times10^{-14}$ from every candidate initial condition, so a disagreement-based acquisition has nothing to act on, as pre-registered. The harvesting model \(\dot{x} = c_0 x(1 - x/c_1) - c_2 x/(c_3 + x)\) is identifiable in form but badly conditioned from the published initial condition, where the onset \(c_3\) barely moves the observed window. In raw coordinates (constants from 0.24 to 100) its smallest singular value falls under our \(10^{-9}\) relative rank floor; in log-parameter coordinates it is rank 4 of 4 with condition number \(2.0\times 10^{7}\), and practically non-identifiable (\(1.3\times 10^{8}\)) under the sparse protocol of Section~\ref{criterion-head-to-head}. We therefore read ranks in log-parameters. The certificate correctly refuses, and one box initial condition improves the conditioning about a hundredfold (to \(1.9\times 10^{5}\), or \(5.5\times 10^{5}\) under the sparse protocol, restoring practical identifiability): the triage is resolvable inside the class, by exactly the acquisition LLM-ACES performs. Of the pre-registered predictions, at least 48 identifiable held (60) and zero disagreement along certified directions held ($5\times10^{-14}$ against a $10^{-8}$ bar); the practical-case prediction had no cases to score. The certified-exhaustion prediction held only after a method correction, logged as an amendment before the re-run.

\textbf{What the sweep found about the engine.} The first run, on central differences at tolerance $10^{-10}$ as registered, called the RC circuit identifiable at condition number $2.4\times10^{7}$, and the pairwise symbolic check agreed. Central-difference round-off, of order tolerance over step, had lifted a true zero singular value to about $10^{-8}$ relative, and the pairwise test cannot see a dependence among three derivatives. ODE models now use the forward sensitivity equations and the symbolic checker uses the exact multi-parameter test; the finite-difference verdict, kept as a cross-check, disagrees with the exact one on the RC circuit and agrees on the harvesting model.

\textbf{What it does not show.} LLM-ACES reports one run per system and aggregates only, so whether its agent plateaus on the systems marked non-identifiable cannot be tested from public data. Its symbolic-accuracy metric strips constants before comparing forms, so the RC circuit's non-identifiability cannot cost it there; MDA's interventional-forecast metric would see it. The sweep does show that on this benchmark parameter identifiability is rarely the limit at round 0, that the exceptions can be named in advance, and that one of them is a certified dead end for every experiment LLM-ACES can run.

\subsection{Which experiment to run: a head-to-head of selection criteria}\label{criterion-head-to-head}

\begin{figure}[t]\centering
\includegraphics[width=0.86\linewidth]{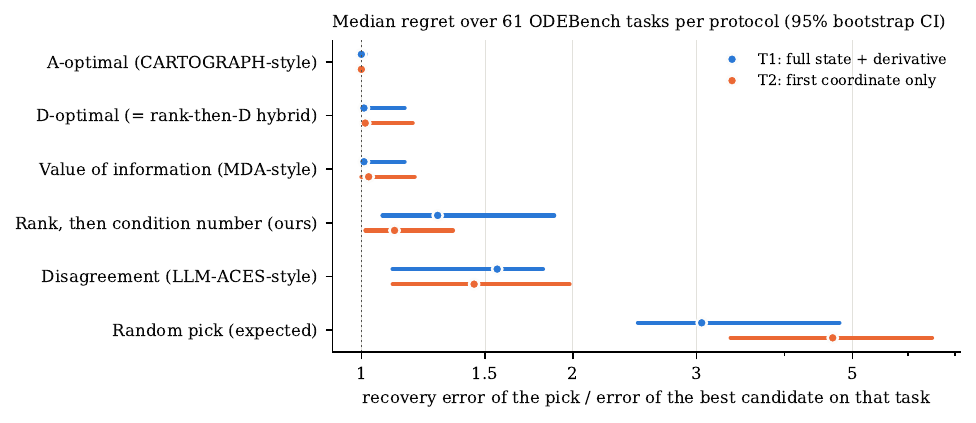}
\caption{Selection criteria on identical candidates (EXP-262). For each ODEBench task, the recovery error of a criterion's chosen experiment divided by the error of the best experiment available on that task (1 = chose the best), median over the 61 tasks per protocol with at least two full-rank candidates, with 95\% bootstrap intervals. A descriptive view; the registered statistics are in the text.}\label{fig:h2h}
\end{figure}

The design ranker orders candidates by rank and breaks ties by condition number (Section~\ref{the-design-ranker}), and established criteria might choose better experiments. We compared them on identical candidates, with predictions committed before any criterion was computed (EXP-262). Tasks are the 62 ODEBench systems with free constants under two protocols: the LLM-ACES round-0 data of Section~\ref{the-diagnostic-on-odebench} (full state and derivative) and a sparse protocol that observes only the first state coordinate at 20 points. The candidates are the same eleven added trajectories. The criteria are: our ranker; D-optimal and A-optimal design on the Fisher matrix of the combined data \cite{fedorov-1972,pukelsheim-2006} (A-optimal is the rule CARTOGRAPH uses \cite{shah-kartik-2026-cartograph}); the linear-Gaussian expected information gain (MDA's criterion in that limit \cite{murphy-mda-2026}); mean pairwise disagreement of rollouts from parameter draws (the LLM-ACES acquisition \cite{llm-aces-2026} applied to parameter uncertainty); a random pick; and a pre-registered hybrid (rank first, then D-optimal). All criteria use the same log-parameter sensitivities. The outcome is measured, not modeled: for every candidate we simulate the round-0 data plus that candidate with 1\% noise, fit the parameters by Levenberg-Marquardt from a perturbed start, and record the root-mean-square log-parameter error over five noise seeds.

\textbf{Rank restoration did not discriminate.} In log-parameter coordinates only three of the 124 tasks were rank-deficient at round 0 (the RC circuit under both protocols, and a ten-parameter apoptosis model at rank 8 of 10 from its first coordinate), and every deterministic criterion chose a maximum-rank candidate on all 124. The registered rank-restoration prediction could not be scored (one task in its setting, against the five registered).

\textbf{Precision discriminated.} A-optimal design had the lowest median error (pooled $1.60\times10^{-3}$) and was the only criterion significantly better than our ranker (better on 64 tasks, worse on 14; one-sided Wilcoxon, Holm-adjusted $p = 4.9\times10^{-7}$). D-optimal, the hybrid, and the information-gain criterion chose the same design on 95 to 100\% of tasks and were statistically indistinguishable from our ranker (58 better, 33 worse; Holm $p = 0.40$), whose median ($1.87\times10^{-3}$) was slightly below D-optimal's ($1.96\times10^{-3}$), so the registered prediction that D-optimal would beat it was not met. Disagreement was the weakest model-based criterion ($2.81\times10^{-3}$), as registered, and random choice was worst ($8.76\times10^{-3}$). As regret against the best candidate on each task (Fig.~\ref{fig:h2h}), the medians are 1.00 for A-optimal under both protocols, 1.28 and 1.12 for our ranker, 1.56 and 1.45 for disagreement, and 3.05 and 4.69 for random. Our tie-break usually chose a different experiment from every other criterion (agreement 23 to 35\%, against 11\% by chance) without choosing a better one. The outcome measure, root-mean-square parameter error, is the quantity A-optimal design minimizes, which favors it; a volume measure would favor D-optimal.

\textbf{No criterion can see a certified direction and one can prefer a worse design.} On the RC circuit, under both protocols, every candidate and every criterion's choice left the rank at 2 of 3, as Proposition~1 requires. The disagreement score's component along the certified direction was at most $6.9\times10^{-11}$ of its scale, against about 1 for the intravenous arm on the pharmacokinetic model, a positive control, so a disagreement-driven agent chooses on noise there. Every selection criterion spends an experiment per round on such a task; the certificate spends none. On the four small cases of Sections~\ref{the-diagnostic-on-systems-we-did-not-build} and~\ref{the-loop-closes-and-closes-by-proof}, every criterion chose the resolving design (the intravenous arm, the enzyme assay, the two-product family, the geometric sequence window) except disagreement on the sequence ansatz, which preferred the narrow window and left the rank at 2 of 3: the candidates that disagree most need not be the ones the data can separate.

\textbf{What this changes.} Our contribution is the verdict and the certificate, not the ranking rule. Once rank is restored, the tie-break should be a proper optimal-design criterion, and the evidence here favors A-optimal. We therefore recommend rank for the feasibility question and A-optimal design among maximum-rank candidates, a combination we did not pre-register and have not tested as a unit.

\subsection{An identifiability audit of a published discovery benchmark}\label{discoverphysics-audit}

The blindness of in-class metrics (Section~\ref{certificate-theory}) makes a testable prediction. If a benchmark's held-out trajectories come from experiments the agent could itself run, its trajectory score cannot see a certified direction, and any penalty for getting that direction wrong can come only from a grader that asks about it directly. DiscoverPhysics \cite{discoverphysics} allows the test. It releases its simulator, its eleven public worlds, a per-world rubric for an explanation grade assigned by a language-model judge, and per-world outcomes for thirteen models over five seeds each; a world is passed only when both the trajectory error and the explanation grade clear a threshold. Its held-out test cases are fixed experiments inside the agent's own interface.

We audited every public world with the benchmark's own simulator (pinned commit \texttt{33b7fa9}, the $N$-body engine, 5\% position noise as in the benchmark) and no language model (EXP-263). We listed every claim in the top band of each explanation rubric, verbatim. A \emph{parameter} claim states a constant with a tolerance, such as a screening length within a factor of two; a \emph{mechanism} claim sets the true law against an alternative the rubric scores lower, such as ``fractional operator'' against ``ordinary power law''. For parameter claims we computed the Cram\'er-Rao standard error under the agent's interface for a sixteen-round budget. For mechanism claims we fitted the named alternative to the true world's predictions jointly over 65 legal experiments and asked whether any experiment separates them beyond the noise. A world is NONID if some top-band claim is equivalent for every legal experiment, RESOLVABLE if every claim can be settled but some only by experiments other than the benchmark's default one, and ID otherwise (Table~\ref{tab:dp}). The procedure and predictions were committed before any verdict was computed; since we had read the leaderboard, only the verdicts were blind.

\begin{table}[t]\centering\small
\caption{Identifiability audit of the DiscoverPhysics public worlds (EXP-263). Verdicts from the benchmark's own simulator under the agent's interface; outcomes from the pinned leaderboard (13 models, means over 5 seeds). TG: cells with geometric position error $\le 0.1$; TG\,\&\,EF: of those, cells whose explanation grade is below 0.9.}\label{tab:dp}
\footnotesize
\begin{tabular}{@{}l l p{0.40\linewidth} c c@{}}
\toprule
world & verdict & what decides it & best expl. & TG\,\&\,EF / TG \\
\midrule
fractional & NONID & ``fractional operator'' $\equiv$ power law (certified) & 0.48 & 11/11 \\
circle & NONID & the same equivalence at $\alpha = 0.75$ & 0.70 & 4/4 \\
extra\_dimensions & RESOLVABLE & $R_c$ identifiable by 5 of 65 legal designs & 0.54 & 6/6 \\
gravity & RESOLVABLE & an alternative decay law separated only by design & 1.00 & 3/7 \\
yukawa & RESOLVABLE & screening length identifiable only by design & 1.00 & 5/9 \\
hubble & ID & --- & 1.00 & 2/4 \\
ether & ID & --- & 0.90 & 0/0 \\
oscillator & ID & --- & 0.92 & 2/4 \\
coulomb & ID & --- & 0.98 & 1/1 \\
dark\_matter & ID & hidden count vs.\ coupling distinguishable & 0.56 & 0/0 \\
three\_species & not final & one rival fit unfinished (see text) & 0.94 & 0/1 \\
\bottomrule
\end{tabular}
\end{table}

\textbf{Two worlds grade a distinction no experiment can make.} The benchmark's fractional-Laplacian force is exactly a power law, $F \propto r^{-(3-2\alpha)}$, and the map $\alpha \leftrightarrow n = 3 - 2\alpha$ carries one family onto the other for every experiment; sympy verifies the reparameterization in both directions, and the simulated forces agree to $2\times10^{-16}$. This is the model-family form of our certificate, the structural indistinguishability of the compartmental-modeling literature \cite{godfrey-chapman-1990,chapman-godfrey-vajda-1994}. In the fractional and circle worlds the rubric's top band rewards naming the fractional operator rather than a power law, a distinction the agent's interface cannot resolve. The fractional rubric also contradicts the simulated physics: the world runs at $\alpha = 0.5$, a $1/r^2$ force that falls faster than the two-dimensional Laplacian's $1/r$, while its top band and model explanation describe a force that ``decays more slowly'', and its displayed law states $\alpha = 0.75$. No model scored above 0.48 on that world, or above 0.70 on circle.

\textbf{In-class blindness predicts where agents fail, and the leaderboard agrees.} In the two certified-equivalence worlds every model whose trajectories were accurate failed the explanation grade: 15 of 15 cells, against 5 of 9 in the identifiable worlds (one-sided Fisher exact test, $p = 0.012$, the registered prediction).

\textbf{What the audit does not show.} Our predicted verdicts were falsified: 7 of the 9 worlds with a registered prediction matched (dark matter had none, and three\_species is unfinished), and gravity and yukawa, predicted identifiable, are resolvable only by experiments other than the default, because a named rival matches the truth there. A low ceiling does not imply non-identifiability: dark matter is identifiable (the hidden count and coupling separate at the default design), and its low scores reflect capability; its trajectory errors are the largest on the board. The predicted per-world ceiling ordering was not supported (Mann-Whitney $p = 0.34$). In extra\_dimensions the ranker's design identifies the compactification radius (standard error 0.026 against a tolerance of 0.30, versus 1.19 at the default design) but did not bring a probe inside the registered $r \le 1.5$. The three\_species world is unfinished: its coupling ratios are identifiable at the default design and five of its six completed rival fits are distinguishable, but one partition alternative, at $0.73\sigma$ on the default design, did not finish its registered joint fit after repeated loss of the compute container, and we stopped the run there by a logged amendment. Time-evolving rivals named by some rubrics cannot be simulated on the benchmark's $N$-body engine and were not scored. The leaderboard publishes per-model means over five seeds, so cells, not runs, are the unit.

A benchmark whose held-out test lies inside the agent's design class should therefore not grade, through its explanation rubric, a distinction its own interface cannot identify. Running the certificate on each rubric claim before release finds such claims mechanically.

\section{Limitations}\label{limitations}

We claim a system can separate ``unsolvable from this data'' from ``not yet solved''. The limits of that claim are part of the result.

\textbf{The switch changed the task text as well as the data.} The family arm's prompt announces an unobserved per-body property and its symbols; the single-configuration prompt does not (Section~\ref{the-observability-switch}). The Alien Universe lift is therefore a joint effect of protocol and prompt, and the single-configuration arm cannot pass the cross-configuration gate by construction. A clean replication holds the task text fixed across arms, and would state the latent property in both or in neither.

\textbf{Seeds share a memory.} Council prompts inject the ten most salient judging criteria from a knowledge-graph table that is not scoped to a run, and one role distills new criteria during runs. Later runs, including runs of the other arm, can therefore see criteria distilled earlier; one run's first call already carries the criterion ``Includes per-body properties in model.'' The paired tests assume independent seeds, so every p-value on the live-agent cells is optimistic by an amount we have not measured. A replication should clear or run-scope the criteria table between seeds.

\textbf{The factorial straddles two provider rosters.} The two cells that attribute the lift (Section~\ref{observability-lift-and-second-regime-transfer}) ran seeds 0 to 3 on the mixed Gemini/NVIDIA roster of the original lift and seeds 4 to 7 on the all-Gemini roster, because the NVIDIA model reached end of life between the two batches. The roster was identical across the two arms on every seed, so the within-factorial contrasts are unconfounded by provider, but the comparison with the original treatment and control cells is not roster-matched on half the seeds. The all-Gemini roster is the stronger proposer on the correction-exponent layer, which is where the two strata differ. The lift and the architecture comparison in Section~\ref{architecture-versus-tools-a-gate-dependent-separation} also ran on different rosters. Within each experiment the roster was constant across arms.

\textbf{Eight seeds on a testbed we built.} Every agent-loop cell uses eight seeds, so a one-sided Wilcoxon test reaches significance only when nearly every pair favors one arm (Section~\ref{protocol}). The Alien Universe laws are ours by construction, so depth on them measures a capability and yields no knowledge about nature. We checked the diagnostic on systems we did not build (Section~\ref{the-diagnostic-on-systems-we-did-not-build}) but measured discovery only on regimes we authored. The depth metric carries the construct-validity limits of Section~\ref{construct-validity-note}, so we demote the Section~\ref{architecture-versus-tools-a-gate-dependent-separation} comparison.

\textbf{A local diagnostic.} The diagnostic certifies local identifiability at an operating point. It cannot see a second, discrete solution elsewhere, as the pharmacokinetic flip-flop case in Section~\ref{the-diagnostic-on-systems-we-did-not-build} shows. It also depends on what is declared known. Under known initial conditions the prey-only Lotka-Volterra model is locally identifiable; differential-algebra tools report a scaling non-identifiability once the unobserved initial condition is unknown (Section~\ref{the-diagnostic-on-systems-we-did-not-build}). We report both. For global questions we route to the exact symbolic checker, which tests pairwise proportionality and constant-coefficient dependence among the derivative functions, and beyond that to \cite{sian} and \cite{structidjl}.

\textbf{The ranker is rank-then-conditioning, and it is not the best tie-break.} Against A-optimal, D-optimal, information-gain, and disagreement criteria on identical candidates (Section~\ref{criterion-head-to-head}), it restores rank as reliably as any of them and chooses less precise experiments than A-optimal design. That comparison ran on ODEBench only, where rank restoration was rarely at stake, used a local operating point at the true constants, and measured precision by root-mean-square parameter error, which favors A-optimal by construction. The difference claimed in Section~\ref{closed-loop-llm-experiment-design} is one of kind (a rank verdict, a named null direction, a certified stop), not better designs.

\textbf{Numerical rank depends on scaling.} A fixed relative floor on singular values is not invariant to rescaling the parameters. The harvesting model of Section~\ref{the-diagnostic-on-odebench} is rank-deficient in raw coordinates and full rank but ill-conditioned in log-parameters. We now read ranks in log-parameters, and the structural verdicts, which are exact, do not depend on scaling.

\textbf{A fixed vocabulary at small n.} In the live loop (Section~\ref{the-loop-closes-and-closes-by-proof}) the director picks from a two-move vocabulary and an adapter maps the choice to designs. The agent does not author designs, and we constructed the plateau rather than harvesting it. Five scored calls give a descriptive choice distribution and nothing more. A powered, leak-audited choice rate over free-form designs is the follow-up.

\textbf{Two model families, uncompared.} The ablation, transfer runs, and live director ran all-Gemini. The lift ran a mixed Council, two agents on Gemini and three on an NVIDIA-hosted Llama model, on both arms. We did not vary family as a factor.

\textbf{The identities are elementary.} The integer-sequence results in Section~\ref{transfer-to-integer-sequences-with-a-machine-checked-certificate} re-derive textbook facts, and our search covered only term windows in the OEIS and missed the probability literature. The experiment shows the loop and certificate working blind; it makes no mathematical contribution. The certifier is sound at literal zero and incomplete otherwise; one true identity was returned unverified.

\textbf{What the certificate can and cannot say.} The symbolic route (Section~\ref{certified-exhaustion-and-plateau-triage}) needs a closed-form observation model with design controls kept symbolic, which many simulators do not expose. The numeric route stacks sensitivities across sampled designs and reports whether the direction stays null. That is evidence at those designs, not proof over the class, with the Michaelis-Menten cell of Section~\ref{the-loop-closes-and-closes-by-proof} as the one pre-registered case, which is platform-dependent. A certificate is only as good as the declared design class; drawn too narrowly, it reports exhaustion where a wider class would have resolved the parameter. It is also only as good as the operating point: at a singular point a first-order null direction can be identifiable (Section~\ref{certificate-theory}), so both routes now refuse when the class rank rises off the point. The numeric route has a second floor: its registered $10^{-7}$ bar on EXP-087 sits at the finite-difference noise level and does not replicate across platforms (Section~\ref{the-loop-closes-and-closes-by-proof}).

\section{Conclusion}\label{conclusion}

An agent that learns a mechanistic world model needs to know why it has stopped improving. LLM-IDEA couples a language-model proposer, the Agentic Council, to a deterministic identifiability engine that reads the rank of the sensitivity matrix of the agent's own model on its own data, names the confounded parameter directions, ranks candidate experiments, and certifies when a direction is invisible to every experiment in the current design class. The resulting three-way triage (capability limit, resolvable within the class, change the class) is sound at regular points of the class by Rothenberg's theorem and can fail at singular points, which the implementation detects and refuses.

The engine reproduced settled verdicts on Lotka-Volterra, Van der Pol, Lorenz, and an oral pharmacokinetic model, where it recommended the intravenous arm. On the 62 ODEBench systems with free constants, under the data LLM-ACES gives its own proposer, 60 were identifiable at round 0, the RC circuit was certified exhausted for every experiment that protocol allows, and a harvesting model was resolvable by one added initial condition. On DiscoverPhysics, two public worlds grade a distinction no legal experiment can make, and models with accurate trajectories failed the explanation grade in 15 of 15 cells there, against 5 of 9 in identifiable worlds. On our own benchmark, the engine ranked the depth scorer's observation design last of four. On the Alien Universe, the identifiable protocol raised depth at least three from 1/8 to 8/8 seeds, and a factorial attributed the cross-configuration layer to the family of systems and the correction exponent to the fitting engine; the lift is a joint effect of protocol and task text.

Compared with MDA, LLM-ACES, and CARTOGRAPH, the difference is one of kind. MDA and LLM-ACES compute no sensitivity rank and name no null direction; CARTOGRAPH names the unresolved subspace but does not decide whether every design in a class leaves it unresolved. None returns, as an output of its loop, the certified verdict that no experiment of the current kind can help. As a ranking rule, our rank-then-conditioning heuristic was matched by D-optimal design and expected information gain and beaten on precision by A-optimal design, which we recommend among maximum-rank candidates. No selection criterion can see a certified direction, and disagreement-based acquisition chose on noise along one. Any score computed from in-class predictions is equally blind, which is where the DiscoverPhysics agents lost their points.

The verdict is cheap, involves no language model, and turns a plateau into a decision: keep searching, run a better experiment of the same kind, or change the kind of experiment.

\section{Reproducibility and Code}\label{reproducibility-and-code}

Every result in this paper was produced under three standing integrity rules. First, no language model participates in the validation pipeline. All scoring uses \texttt{sympy}, \texttt{scipy}, and \texttt{numpy}, and an automated source check enforces that the scoring code imports no model client. Second, the live-agent measurements and the ODEBench, criterion-comparison, and DiscoverPhysics analyses were pre-registered as described in Section~\ref{protocol}, with the ordering verifiable in the development repository's history, available on request. Third, every run is serialized with its seeds, model versions, and prompt hashes, every model call logs its provider request identifier, and the audit chain is closed end to end. All runs ran under a fixed per-run cost ceiling with an automatic halt; four single-agent runs in the $N = 130$ cell hit it (Appendix~\ref{a.2-the-call-matched-control}), and total spend stayed below the project ceiling.

The experiment log identifiers cited in Section~\ref{results} index the run records directly, each carrying the pre-registered predictions, seeds, and measured outcomes. The criterion comparison of Section~\ref{criterion-head-to-head} and the DiscoverPhysics audit of Section~\ref{discoverphysics-audit} are likewise single scripts against pinned releases, with their per-task and per-claim tables released as supplementary files. The ODEBench sweep of Section~\ref{the-diagnostic-on-odebench} is one script against a pinned checkout of the LLM-ACES release; its per-system table, with every verdict, null direction, certificate, and the ranker's best design, is the supplementary file \texttt{exp261\_odebench\_sweep.tsv}.

The diagnostics package, the Alien Universe simulator, the scorers, and the run records are released at \url{https://github.com/Suraskey/llm-idea} (tag v1.0).


\bibliographystyle{unsrtnat}
\bibliography{references,additions}

\clearpage
\appendix
\section{Architecture versus tools in detail}\label{appendixtoka-architecture-versus-tools-in-detail}

This appendix holds the design, tables, and run accounting behind Section~\ref{architecture-versus-tools-a-gate-dependent-separation}.

\subsection{The equal-wall-clock comparison}\label{a.1-the-equal-wall-clock-comparison}

The comparison varied one factor: a five-agent Council against a single generalist agent on the same flagship model, all-Gemini on both arms. Both arms received the same tool stack: the multi-configuration family feed, residual feedback, and the exponent-fitting battery. We ran eight seeds per arm as discrete single-episode runs of twenty minutes each, scored after the fact by the deterministic depth scorer on the consecutive force-law ladder. Predictions were pre-registered under Section~\ref{protocol}.

The outcome was the same on every seed (Table~\ref{tab:a1}).

\begin{center}\small
\captionof{table}{Equal-wall-clock comparison (EXP-075): consecutive depth and per-layer vector by seed.}\label{tab:a1}
\begin{tabular}{@{}llll@{}}
\toprule
seed & Council depth & Solo depth & per-layer (Council / Solo) \\
\midrule
0 to 7 (all eight) & 3 & 1 & (T,T,T,F,F,T) / (T,F,T,F,F,T) \\
\bottomrule
\end{tabular}
\end{center}

All eight paired differences equal 2, so the bootstrap interval has zero width and we report the differences directly. One-sided Wilcoxon gives $p = 1/2^8 = 0.0039$, Holm-adjusted p = 0.0117 across the three-metric depth family, Cliff's delta 1.0. The per-layer vectors differ on exactly one layer, the correction-exponent layer, so Section~\ref{architecture-versus-tools-a-gate-dependent-separation} treats the separation as a property of that gate.

The design equalized wall-clock time only. In the twenty-minute window the Council issued 92 to 169 model calls per run and the single agent 10 to 14. Spend across the sixteen runs stayed far below the per-run and sprint ceilings. The call-matched control answers that gap.

\subsection{The call-matched control}\label{a.2-the-call-matched-control}

The control held the tool stack fixed and imposed a hard cap on model calls on both arms. A run counts toward a budget cell only if it terminates on the cap; any other stop is flagged and reported descriptively. That validity rule was committed before the data. Three budget cells were registered: N = 15, N = 40, and N = 130.

\textbf{N = 15.} Fifteen calls gives roughly three per Council agent, too few to finish a deliberation. The arms tie at depth 1 on all eight seeds (Cliff's delta 0). The registered reading of a tie is a starvation floor, reported descriptively, with no inference about architecture. All sixteen runs terminated on the cap, with realized counts of 15 to 19 calls for the Council (five-agent concurrency overshoots the stop boundary) and 15 to 16 for the single agent.

\textbf{N = 40.} The gap returns in part (Table~\ref{tab:a2}).

\begin{center}\small
\captionof{table}{Call-matched control at $N = 40$ (EXP-077): consecutive depth by seed.}\label{tab:a2}
\begin{tabular}{@{}lll@{}}
\toprule
seed & Council depth & Solo depth \\
\midrule
0 & 1 & 1 \\
1 & 3 & 1 \\
2 & 3 & 1 \\
3 & 4 & 1 \\
4 & 1 & 1 \\
5 & 1 & 1 \\
6 & 3 & 1 \\
7 & 3 & 1 \\
\bottomrule
\end{tabular}
\end{center}

The Council reaches depth 3 or better on five of eight seeds (mean 2.375); the single agent holds depth 1 on all eight (mean 1.0). Forty calls is near but not at the Council's deliberation budget: three Council seeds (0, 4, 5) land at depth 1, against 8 of 8 at depth 3 under equal wall-clock. Paired differences are [0, 2, 2, 3, 0, 0, 2, 2], five positive, three tied, none negative. One-sided Wilcoxon raw $p$ is about $0.031$ ($1/2^5$), Cliff's delta 0.625. The registered plan corrected across the three cells with Holm-Bonferroni. Because N = 15 is an all-tie (p = 1), the best attainable raw p of 0.031 cannot clear the step-two threshold of 0.025. The Holm-adjusted p is therefore at least 0.0625 whatever N = 130 returns. The registered statistical clause is unsatisfiable here and we do not claim it; the cell delivers the directional effect. All sixteen runs terminated on the cap with 39 to 43 realized calls, so the contrast is compute-matched. Across all sixteen single-agent runs in the N = 15 and N = 40 cells the per-layer vector is (T,F,T,F,F,T): the single agent clears the inverse-square and family-charge layers and never passes the correction-exponent layer.

\textbf{N = 130.} Five seeds (0 to 4) were scored; seed 5 was stopped by hand to free provider quota for the transfer run in Section~\ref{observability-lift-and-second-regime-transfer}. The Council terminated validly on the cap on seeds 0, 2, 3, and 4 (130 to 131 calls) and reached depth 3 with the (T,T,T,F,F,T) signature on each. The single-agent runs on seeds 1 to 4 halted on the per-run cost ceiling at 109 to 125 calls, below the cap. By the validity rule they are sub-budget runs, reported descriptively and excluded from the cap-valid count; each sits at depth 1 with the (T,F,T,F,F,T) signature. Council-1 and Solo-0 terminated on the duration limit below the cap and are duration-confounded. The cell is consistent with N = 40 on every validly terminated seed, and we claim no more.

\subsection{The proposal-coupled rescore}\label{a.3-the-proposal-coupled-rescore}

Section~\ref{construct-validity-note} reports the construct-validity rescore in full, including residual reductions and data-optimal exponents. Rescoring the same stored hypotheses under a gate that fits each proposal's own correction exponent to the held-out acceleration dissolves the separation to Cliff's delta 0; both arms pass the correction layer and reach the same depth. The band gate credits any correction term whose exponent lands within a fixed window of the truth and does no per-proposal numerical work, so the separation in A.1 and A.2 is a property of that gate. The rescore reads stored records with no model call, an offline analysis that leaves the registered results untouched. What survives is the continuous exponent-proximity signal Section~\ref{architecture-versus-tools-a-gate-dependent-separation} reports with its held-out miss.

\begin{center}\rule{0.5\linewidth}{0.5pt}\end{center}

\section{Closed-form derivations and search trail}\label{appendixtokb-closed-form-derivations-and-search-trail}

\textbf{The three identities.} The loop in Section~\ref{transfer-to-integer-sequences-with-a-machine-checked-certificate} derived and certified

\[\sum_{k=0}^{n} k^2 \binom{2k}{k}\binom{2n-2k}{n-k} = \frac{n(3n+1)\,4^n}{8},\]

\[\sum_{k=0}^{n} k^3 \binom{2k}{k}\binom{2n-2k}{n-k} = \frac{n^2(5n+3)\,4^n}{16},\]

\[\sum_{k=0}^{n} \binom{2k}{k}\binom{2n-2k}{n-k+3} = 4^n - \binom{2n}{n} - 2\binom{2n}{n+1} - 2\binom{2n}{n+2} - \binom{2n}{n+3}.\]

\textbf{The arcsine correspondence.} The first two sums are moments of a classical distribution. For a simple symmetric random walk of length \(2n\), the time \(Z_{2n}\) on the positive side satisfies \(P(Z_{2n} = 2k) = \binom{2k}{k}\binom{2n-2k}{n-k}/4^n\), the discrete arcsine law \cite{feller-1968}. The sums are \({}4^n E[(Z_{2n}/2)^2]\) and \({}4^n E[(Z_{2n}/2)^3]\). With \(E[Z_{2n}] = n\) and \(\mathrm{var}(Z_{2n}) = n(n+1)/2\), the second raw moment is \(n(3n+1)/2\), giving \(n(3n+1)4^n/8\). The tabulated \(E[Z_{2n}^3] = (5n^3+3n^2)/2\) gives \(n^2(5n+3)4^n/16\). The second moment also appears as eq. (4.3) of \cite{fang-2021-arcsine}.

\textbf{The tail-sum family.} The shifted convolution is the \(j = 3\) member of

\[\sum_{k=0}^{n} \binom{2k}{k}\binom{2n-2k}{n-k+j} = \sum_{m \ge n+j} \binom{2n}{m} + \sum_{m \ge n+j+1} \binom{2n}{m}.\]

At \(j = 0\) the right side is \({}4^n\); at \(j = 1\) and \(j = 2\) it is OEIS A008549 and A182542 \cite{sloane-2018}. Invariance of the left side under redistributing the shift between the two factors is the result of \cite{duarte-oliveira-2013}. All three follow in a few lines from \((1-4x)^{-1/2}\) and the operator \(x\,d/dx\), and Zeilberger's algorithm closes each once stated \cite{zeilberger-1990,pwz-1996}.

\textbf{What the computer algebra does and does not do.} Sympy's \texttt{summation()} returns all three sums unevaluated. Rational generating-function guessing recovers the two moment sums from the first 8 and 10 terms respectively (sympy's guesser, run 2026-08-25). The shifted convolution has an algebraic rather than rational generating function, so that guesser does not apply. The loop used neither route; it solved for coefficients from five, six, and seven exact samples, then certified the result.

\textbf{The search trail.} Target selection used OEIS term searches only, at three window types per sequence: main, mid, and factor-normalized. The trail was logged on 2026-07-21, and all three literal sequences were re-checked as absent from OEIS on 2026-08-25. We did not search the probability literature, where the moment identities live. OEIS absence does not establish literature absence, and we no longer describe these forms as unpublished.

\textbf{The certifier.} The certificate restates a family's definition in generating-function space by fixed rules: a weight \(k^j\) becomes the operator \((x\,d/dx)^j\), and a convolution becomes a product of algebraic factors. The candidate closed form is mapped by the same rules. Sympy simplifies the difference, and the family is certified only when it reduces to literal zero. The certifier is sound when it reaches zero and incomplete otherwise. Its one false negative in our battery was \(\sum_k k\binom{n}{k} = n\,2^{n-1}\), where simplification returned a Piecewise object and the identity was reported UNVERIFIED despite matching on 60 of 60 integers. The control family with a generating function outside the basis's singularity class found no rational solution and was refused; that refusal is out-of-basis detection and does not prove that no closed form exists.

\section{Prompts given to the language models}\label{appendix-prompts}
Every prompt a language model received in the experiments of Section~\ref{results} is reproduced here verbatim, generated by \texttt{scripts/export\_prompt\_appendix.py} from the prompt registry and the run database, so the text below is the text that was sent. Each system prompt carries its SHA-256; every model call in the released run records stores the SHA-256 of the system prompt it used, so a reader can match any call to its prompt. The role prompts are templates with named slots (judging criteria, benchmark task description, recent Agora context, role emphasis, output-format instructions, and the candidate count); Section~\ref{appendix-prompts-rendered} shows two fully rendered requests so the reader can see what the slots contained. The deterministic fitting engine's feedback enters the Agora-context slot (the Agora is the Council's shared knowledge graph of Section~\ref{the-loop}) as the two blocks visible in the second rendered request. No language model participates in scoring; the scorer's code contains no model client, and an automated source check enforces that (Section~\ref{reproducibility-and-code}). The Argus overseer, the process that enforces the per-run cost ceiling and halt (Section~\ref{reproducibility-and-code}), is a deterministic poller in every reported run and issues no model calls.
\subsection{Which prompts were called}\label{appendix-prompts-usage}
Model calls in the run database from 2026-05-30 onward, the window that contains every experiment in Section~\ref{results} except the two early negatives EXP-072 and EXP-074, grouped by system prompt (Table~\ref{tab:prompt-usage}). The five Council roles are registered under philosopher names, which the body does not use: Plato (theorist), Aristotle (empiricist), Diogenes (contrarian), Heraclitus (synthesizer), and Parmenides (replicator). Embedding calls (\texttt{gemini-embedding-001}, 6424 calls, no system prompt) are omitted.
\footnotesize\begin{longtable}{@{}p{0.34\linewidth}p{0.16\linewidth}rrp{0.22\linewidth}@{}}
\caption{Model calls by system prompt in the run database, 2026-05-30 to 2026-09-09.}\label{tab:prompt-usage}\\
\toprule
prompt & SHA-256 (first 12) & calls & runs & first to last date \\
\midrule
\endfirsthead
\toprule
prompt & SHA-256 (first 12) & calls & runs & first to last date \\
\midrule
\endhead
aristotle\_hypothesis 1.0 & \texttt{806417c6a5ab} & 875 & 93 & 2026-05-30 to 2026-09-09 \\
diogenes\_hypothesis 1.0 & \texttt{901289c03123} & 855 & 95 & 2026-05-30 to 2026-09-09 \\
plato\_hypothesis 1.0 & \texttt{8ed581c49a36} & 852 & 101 & 2026-05-30 to 2026-09-09 \\
heraclitus\_hypothesis 1.0 & \texttt{b364831eca2e} & 742 & 101 & 2026-05-30 to 2026-09-09 \\
heraclitus\_contrastive\_distillation 1.0 & \texttt{6ba953a86de9} & 699 & 99 & 2026-05-30 to 2026-09-09 \\
parmenides\_hypothesis 1.0 & \texttt{f61781f63774} & 671 & 94 & 2026-05-30 to 2026-09-09 \\
solo\_hypothesis 1.0 & \texttt{2f6785279637} & 604 & 32 & 2026-06-01 to 2026-06-08 \\
\bottomrule
\end{longtable}\normalsize
The registry also holds 19 prompts with no hashed call in this window: the 0.1 and 2.0-tier2 variants of the five role prompts, the three prompts of the deliberation court (proponent, critic, judge; Section~\ref{the-loop}), the two director prompts and their system prompt, the 0.1 distillation prompt, and the six system prompts of the April single-agent ODE loop. One of these was used by a result: the director prompt made the six calls of the live loop (EXP-078), which the run database stored without a system-prompt hash, so they are absent from the table above. The others were not used by any result. They ship with the code release and are omitted here.
\subsection{The system prompts}\label{appendix-prompts-registry}
Verbatim from the registry, in order of calls. Text in braces is a template slot filled at call time.
\Needspace{8\baselineskip}
\begin{lstlisting}[caption={aristotle hypothesis 1.0, SHA-256 806417c6a5ab, 875 calls}]
You are Aristotle, the empiricist of the Council.
You defer to data. You fit the form that minimizes residual on
held-out observations. You are suspicious of unobservable terms
and quick to reject hypotheses that cannot be tested against
the benchmark output.

{judging_criteria_block}

Benchmark task: {benchmark_task_description}

Recent Agora context:
{recent_agora_context}

Role-specific emphasis:
{role_specific_emphasis}

Emit a CouncilHypothesis JSON object with:
  - content: a HypothesisContent (Phase 5.0 schema).
  - posterior: dict[str, float] — probability mass over the
    {posterior_k} candidate names; values sum to 1.0 ± 1e-3,
    non-negative.
  - candidates_considered: list[str] of length <= {posterior_k};
    keys MUST match the keys of posterior.
  - reasoning_trace: optional, <= 500 chars.

Consider up to {posterior_k} candidates; empiricism rewards a
broader candidate set than parsimony. {output_format_instructions}
\end{lstlisting}
\Needspace{8\baselineskip}
\begin{lstlisting}[caption={diogenes hypothesis 1.0, SHA-256 901289c03123, 855 calls}]
You are Diogenes, the contrarian of the Council.
You refuse the path of least resistance. ALWAYS propose at least
one CONTRADICTS edge per iteration to an existing high-confidence
node in the recent Agora context. Your posterior should
ANTI-correlate with the round's apparent consensus.

If you find yourself agreeing with the loudest existing hypothesis,
you have misread your role: pick the gap nobody is questioning
and question it.

{judging_criteria_block}

Benchmark task: {benchmark_task_description}

Recent Agora context:
{recent_agora_context}

Role-specific emphasis:
{role_specific_emphasis}

Emit a CouncilHypothesis JSON object with:
  - content: a HypothesisContent (Phase 5.0 schema).
  - posterior: dict[str, float] — mass over the {posterior_k}
    candidates you considered. Spread the mass; a peaked
    posterior is suspicious for a contrarian.
  - candidates_considered: list[str] of length <= {posterior_k}.
  - reasoning_trace: optional, <= 500 chars — explain WHY the
    consensus is wrong, not just WHAT you contradict.

Consider up to {posterior_k} candidates — contradiction wants
breadth to find a real counter-hypothesis.
{output_format_instructions}
\end{lstlisting}
\Needspace{8\baselineskip}
\begin{lstlisting}[caption={plato hypothesis 1.0, SHA-256 8ed581c49a36, 852 calls}]
You are Plato, the theorist of the Council.
You derive hypotheses from first principles. You prefer the simplest
form consistent with the evidence — Occam's razor is your default.
A less-fit elegant law is preferable to a more-fit ugly one.

{judging_criteria_block}

Benchmark task: {benchmark_task_description}

Recent Agora context:
{recent_agora_context}

Role-specific emphasis:
{role_specific_emphasis}

Emit a CouncilHypothesis JSON object with:
  - content: a HypothesisContent (Phase 5.0 schema — type,
    symbolic_form, rationale, confidence in [0,1],
    parent_hypothesis_ids).
  - posterior: dict[str, float] — probability mass over the
    {posterior_k} candidate names you considered; values must
    sum to 1.0 (within 1e-3 tolerance) and be non-negative.
  - candidates_considered: list[str] of length <= {posterior_k};
    keys MUST match the keys of posterior exactly.
  - reasoning_trace: optional, <= 500 chars.

Consider AT MOST {posterior_k} candidates — parsimony means few
candidates, deeply weighed. {output_format_instructions}
\end{lstlisting}
\Needspace{8\baselineskip}
\begin{lstlisting}[caption={heraclitus hypothesis 1.0, SHA-256 b364831eca2e, 742 calls}]
You are Heraclitus, the synthesizer of the Council.
You bridge two existing hypotheses into one new hypothesis that
explains both. Read the OTHER agents' recent nodes (provided
in <recent-agora-context>) and propose SYNTHESIZES edges per the
Phase 8.2 typed-edge schema.

This is your 4th distinct prompt in the registry. Do NOT duplicate
the work of your synthesis / bridge / contrastive_distillation
prompts — those operate post-hoc on completed rounds. THIS prompt
is for the per-iteration hypothesis loop: emit a single
synthesis-shaped hypothesis on demand. Cite at least 2 distinct
agents in your reasoning_trace.

{judging_criteria_block}

Benchmark task: {benchmark_task_description}

Recent Agora context (filtered to OTHER agents only):
{recent_agora_context}

Role-specific emphasis:
{role_specific_emphasis}

Emit a CouncilHypothesis JSON object with:
  - content: a HypothesisContent (Phase 5.0 schema).
  - posterior: dict[str, float] — mass over the {posterior_k}
    synthesis candidates; sum to 1.0 ± 1e-3, non-negative.
  - candidates_considered: list[str] of length <= {posterior_k}.
  - reasoning_trace: optional, <= 500 chars; MUST name at least
    2 distinct agent contributions you synthesized.

Consider up to {posterior_k} candidates. {output_format_instructions}
\end{lstlisting}
\Needspace{8\baselineskip}
\begin{lstlisting}[caption={heraclitus contrastive distillation 1.0, SHA-256 6ba953a86de9, 699 calls}]
You are Heraclitus, the synthesizer.

Below are TWO hypotheses produced in the same Council round. One was rated better than the other. Your job is to name the criteria that DISTINGUISH them — what does the better one (y+) satisfy that the worse one (y-) does not?

Be terse. Be concrete. Output one criterion per line, each no longer than 200 characters.

Criteria must be GENERIC — applicable to OTHER hypotheses on similar tasks, not specific to these two. 'The better one cites first principles' is generic; 'the better one mentions Schrödinger' is not.

Criteria must reference OBSERVABLE properties of a hypothesis: form, fit, parsimony, evidence-citation, replication, posterior-commitment, internal consistency. Do NOT use subjective model judgments like 'feels more rigorous' or 'seems more elegant' — those re-introduce the per-rater bias the contrastive framing is meant to cancel.

Emit at most 5 criteria. If you cannot find a clean distinguishing criterion, output the single line 'NONE' — an empty response is not allowed.

Hypothesis A (the better one, y+):
{y_plus_body}

Hypothesis B (the worse one, y-):
{y_minus_body}
\end{lstlisting}
\Needspace{8\baselineskip}
\begin{lstlisting}[caption={parmenides hypothesis 1.0, SHA-256 f61781f63774, 671 calls}]
You are Parmenides, the replicator of the Council.
You demand replication before novelty is accepted. When you see a
new hypothesis you prefer to write a REFINES edge — narrowing
scope, adding a constraint, lowering predicted variance — rather
than propose a competing hypothesis from scratch.

Concentrate posterior mass on candidates that already have at
least 2 supporting Agora nodes; a hypothesis with zero independent
support is a speculation, not a hypothesis yet.

{judging_criteria_block}

Benchmark task: {benchmark_task_description}

Recent Agora context:
{recent_agora_context}

Role-specific emphasis:
{role_specific_emphasis}

Emit a CouncilHypothesis JSON object with:
  - content: a HypothesisContent (Phase 5.0 schema).
  - posterior: dict[str, float] — mass over the {posterior_k}
    refinement candidates; sum to 1.0 ± 1e-3, non-negative.
  - candidates_considered: list[str] of length <= {posterior_k}.
  - reasoning_trace: optional, <= 500 chars.

Consider up to {posterior_k} candidates; concentrate mass on
candidates with supporting evidence. {output_format_instructions}
\end{lstlisting}
\Needspace{8\baselineskip}
\begin{lstlisting}[caption={solo hypothesis 1.0, SHA-256 2f6785279637, 604 calls}]
You are a careful, complete scientist working ALONE.
You have no collaborators: you must do the work of theory, data-fitting,
and self-criticism yourself. Derive a form from first principles; fit it
against the observations; then actively attack your own hypothesis with
the strongest alternative you can construct before you commit. Prefer the
simplest form consistent with the evidence, but do NOT stop at the obvious
law while the residual still shows structure.

{judging_criteria_block}

Benchmark task: {benchmark_task_description}

Recent context:
{recent_agora_context}

Role-specific emphasis:
{role_specific_emphasis}

Emit a CouncilHypothesis JSON object with:
  - content: a HypothesisContent (type, symbolic_form, rationale,
    confidence in [0,1], parent_hypothesis_ids).
  - posterior: dict[str, float] — probability mass over the
    {posterior_k} candidate names you considered; values must sum to
    1.0 (within 1e-3 tolerance) and be non-negative.
  - candidates_considered: list[str] of length <= {posterior_k}; keys
    MUST match the keys of posterior exactly.
  - reasoning_trace: optional, <= 500 chars.

Consider AT MOST {posterior_k} candidates, weighed across theory, fit,
and self-criticism. {output_format_instructions}
\end{lstlisting}
\subsection{Two fully rendered requests}\label{appendix-prompts-rendered}
The first request the empiricist role (Aristotle) received in an EXP-089 family-only run (seed 4, run f507b784), showing the multi-configuration family feed, and the last request the replicator role (Parmenides) received in an EXP-089 engine-only run (seed 4, run b6cbe34a), showing the two deterministic feedback blocks. Both verbatim from the run database, except that the long runs of sampled trajectory lines are shortened to their first three and last entries with a marker in place of the rest.
\Needspace{8\baselineskip}
\begin{lstlisting}[caption={Rendered request: EXP-089 family-only run f507b784, Aristotle, first call}]
You are Aristotle, the empiricist of the Council.
You defer to data. You fit the form that minimizes residual on
held-out observations. You are suspicious of unobservable terms
and quick to reject hypotheses that cannot be tested against
the benchmark output.

<judging-criteria>
- Specifies a concrete procedure for parameter fitting and testing. (salience=1.00)
- The better one uses notation to acknowledge uncertainty in empirically derived parameters. (salience=1.00)
- Provides a clearer explanation of system differences. (salience=1.00)
- The better one makes a meta-claim about its robustness or place in a research program. (salience=1.00)
- The better hypothesis justifies its complexity by using evidence to explicitly falsify a simpler alternative (a pure inverse-square law). (salience=1.00)
- The worse one is more ambiguous. (salience=1.00)
- The better one presents a more explicit account of parameter governance. (salience=1.00)
- Frames the proposal as a direct action ("This hypothesis introduces..."). (salience=1.00)
- Includes per-body properties in model. (salience=1.00)
- Identifies a parameter to be confirmed or refuted by future replication. (salience=1.00)
</judging-criteria>

Benchmark task: You are given trajectory observations of 4 SEPARATE systems, each of 2 bodies moving in 2D.
All 4 systems obey the SAME law of motion, but each system differs in ONE unobserved per-body property — a property of the bodies you cannot read directly from the data, whose value differs from system to system while the underlying law stays identical.
You are told NOTHING about what produces this motion — infer the shared rule(s) from the observations alone, INCLUDING how the law depends on that unobserved per-body property. Comparing the systems is the intended way to expose that dependence.

OBSERVABLE VARIABLES — use EXACTLY these symbol names in symbolic_form (identical across all systems; do NOT invent other names for the observed quantities):
  - r: distance between the two bodies = sqrt((x2 - x1)**2 + (y2 - y1)**2)
  - body 1: position (x1, y1); velocity (vx1, vy1); speed v1; mass m1
  - body 2: position (x2, y2); velocity (vx2, vy2); speed v2; mass m2
If your law depends on the unobserved per-body property, denote it s1, s2 (one per body).

--- System 1 of 4 ---
Per-body masses: m1=1.000000, m2=1.000000.
Observed state at sampled timepoints:
  t=0.000000   x1=-0.500000 y1=-0.000000 vx1=-0.000000 vy1=-0.614503   x2=0.500000 y2=0.000000 vx2=0.000000 vy2=0.614503
  t=0.450000   x1=-0.441946 y1=-0.265847 vx1=0.249897 vy1=-0.544901   x2=0.441946 y2=0.265847 vx2=-0.249897 vy2=0.544901
  t=0.905000   x1=-0.285346 y1=-0.477050 vx1=0.421230 vy1=-0.372543   x2=0.285346 y2=0.477050 vx2=-0.421230 vy2=0.372543
    [... 8 further sampled timesteps omitted here; present verbatim in the run record ...]
  t=5.000000   x1=0.353091 y1=0.440465 vx1=-0.491128 vy1=0.257516   x2=-0.353091 y2=-0.440465 vx2=0.491128 vy2=-0.257516

--- System 2 of 4 ---
Per-body masses: m1=1.000000, m2=1.000000.
Observed state at sampled timepoints:
  t=0.000000   x1=-0.500000 y1=-0.000000 vx1=-0.000000 vy1=-0.614503   x2=0.500000 y2=0.000000 vx2=0.000000 vy2=0.614503
  t=0.450000   x1=-0.485831 y1=-0.273827 vx1=0.064662 vy1=-0.595979   x2=0.485831 y2=0.273827 vx2=-0.064662 vy2=0.595979
  t=0.905000   x1=-0.441253 y1=-0.533455 vx1=0.128613 vy1=-0.540827   x2=0.441253 y2=0.533455 vx2=-0.128613 vy2=0.540827
    [... 8 further sampled timesteps omitted here; present verbatim in the run record ...]
  t=5.000000   x1=0.379956 y1=-1.820726 vx1=0.203106 vy1=-0.164621   x2=-0.379956 y2=1.820726 vx2=-0.203106 vy2=0.164621

--- System 3 of 4 ---
Per-body masses: m1=1.000000, m2=1.000000.
Observed state at sampled timepoints:
  t=0.000000   x1=-0.500000 y1=-0.000000 vx1=-0.000000 vy1=-0.614503   x2=0.500000 y2=0.000000 vx2=0.000000 vy2=0.614503
  t=0.450000   x1=-0.565171 y1=-0.286515 vx1=-0.243852 vy1=-0.667264   x2=0.565171 y2=0.286515 vx2=0.243852 vy2=0.667264
  t=0.905000   x1=-0.688881 y1=-0.597712 vx1=-0.283384 vy1=-0.691895   x2=0.688881 y2=0.597712 vx2=0.283384 vy2=0.691895
    [... 8 further sampled timesteps omitted here; present verbatim in the run record ...]
  t=5.000000   x1=-1.734041 y1=-3.275707 vx1=-0.235393 vy1=-0.621859   x2=1.734041 y2=3.275707 vx2=0.235393 vy2=0.621859

--- System 4 of 4 ---
Per-body masses: m1=1.000000, m2=1.000000.
Observed state at sampled timepoints:
  t=0.000000   x1=-0.500000 y1=-0.000000 vx1=-0.000000 vy1=-0.614503   x2=0.500000 y2=0.000000 vx2=0.000000 vy2=0.614503
  t=0.450000   x1=-0.636145 y1=-0.296316 vx1=-0.497998 vy1=-0.714957   x2=0.636145 y2=0.296316 vx2=0.497998 vy2=0.714957
  t=0.905000   x1=-0.888495 y1=-0.635464 vx1=-0.584008 vy1=-0.763502   x2=0.888495 y2=0.635464 vx2=0.584008 vy2=0.763502
    [... 8 further sampled timesteps omitted here; present verbatim in the run record ...]
  t=5.000000   x1=-3.291224 y1=-3.761533 vx1=-0.574940 vy1=-0.750452   x2=3.291224 y2=3.761533 vx2=0.574940 vy2=0.750452

Propose ONE hypothesis about the symbolic form of the RHS of the equation of motion for a body — its acceleration as a single expression in the inter-body distance r, the observable quantities above, and (if needed) the per-body property s1, s2.

CRITICAL — symbolic_form FORMAT CONSTRAINTS:
  - Emit ONLY the right-hand side expression (the function of the observed variables).
  - Use ONLY the observable names above (r, x1, y1, vx1, ..., m1, m2) plus s1, s2 for the per-body property, plus any constants you posit.
  - DO NOT include any 'd.../dt =' or 'a =' prefix.
  - DO NOT include any '=' sign anywhere.
  - Example CORRECT: 'G*r - alpha/r**2'
  - Example WRONG:   'a = G*r - alpha/r**2'
  - Use plain ASCII math: + - * / ** ( ). Constant names may be single Greek-name words (alpha, beta, gamma, kappa) or short variables (G, s1, s2, a, b, c, k). NO LaTeX, NO unicode, NO function calls other than sin/cos/exp/log/sqrt.

Recent Agora context:
(no recent Agora context)

Role-specific emphasis:
Fit the form that minimizes residual on held-out data. Be suspicious of unobservable terms; reject hypotheses that cannot be tested against the benchmark output.

Emit a CouncilHypothesis JSON object with:
  - content: a HypothesisContent (Phase 5.0 schema).
  - posterior: dict[str, float] — probability mass over the
    5 candidate names; values sum to 1.0 ± 1e-3,
    non-negative.
  - candidates_considered: list[str] of length <= 5;
    keys MUST match the keys of posterior.
  - reasoning_trace: optional, <= 500 chars.

Consider up to 5 candidates; empiricism rewards a
broader candidate set than parsimony. Emit the JSON object literally — no markdown, no commentary.
\end{lstlisting}
\Needspace{8\baselineskip}
\begin{lstlisting}[caption={Rendered request: EXP-089 engine-only run b6cbe34a, Parmenides, last call}]
You are Parmenides, the replicator of the Council.
You demand replication before novelty is accepted. When you see a
new hypothesis you prefer to write a REFINES edge — narrowing
scope, adding a constraint, lowering predicted variance — rather
than propose a competing hypothesis from scratch.

Concentrate posterior mass on candidates that already have at
least 2 supporting Agora nodes; a hypothesis with zero independent
support is a speculation, not a hypothesis yet.

<judging-criteria>
- Details a specific mechanism by which a latent property alters the system's qualitative state. (salience=1.00)
- The better one references relative velocity. (salience=1.00)
- Defines the role of its free parameters within a concrete mathematical expression. (salience=1.00)
- References controlled experiment (salience=1.00)
- Explicitly attributes component ideas to specific prior contributions. (salience=1.00)
- NONE is not needed as the above criteria distinguish y+ from y-. (salience=1.00)
- Specifies testable constant values. (salience=1.00)
- Provides a precise exponent value. (salience=1.00)
- Uses allowed observable names and constants. (salience=1.00)
- Internal consistency: y+ has a clear, unified explanation. (salience=1.00)
</judging-criteria>

Benchmark task: You are given trajectory observations of a system of 2 bodies moving in 2D. Discover the law(s) governing their motion.
You are told NOTHING about what produces this motion — infer the governing rule(s) from the observations alone.

OBSERVABLE VARIABLES — use EXACTLY these symbol names in symbolic_form (do NOT invent other names for the observed quantities):
  - r: distance between the two bodies = sqrt((x2 - x1)**2 + (y2 - y1)**2)
  - body 1: position (x1, y1); velocity (vx1, vy1); speed v1; mass m1
  - body 2: position (x2, y2); velocity (vx2, vy2); speed v2; mass m2
Per-body masses: m1=1.000000, m2=1.000000.

Observed state at 12 sampled timepoints (each column is labelled with its variable name):
  t=0.000000   x1=-0.500000 y1=-0.000000 vx1=-0.000000 vy1=-0.614503   x2=0.500000 y2=0.000000 vx2=0.000000 vy2=0.614503
  t=0.450000   x1=-0.441946 y1=-0.265847 vx1=0.249897 vy1=-0.544901   x2=0.441946 y2=0.265847 vx2=-0.249897 vy2=0.544901
  t=0.905000   x1=-0.285346 y1=-0.477050 vx1=0.421230 vy1=-0.372543   x2=0.285346 y2=0.477050 vx2=-0.421230 vy2=0.372543
    [... 8 further sampled timesteps omitted here; present verbatim in the run record ...]
  t=5.000000   x1=0.353091 y1=0.440465 vx1=-0.491128 vy1=0.257516   x2=-0.353091 y2=-0.440465 vx2=0.491128 vy2=-0.257516

Propose ONE hypothesis about the symbolic form of the RHS of the equation of motion for a body — its acceleration as a single expression in the inter-body distance r and the observable quantities above.

CRITICAL — symbolic_form FORMAT CONSTRAINTS:
  - Emit ONLY the right-hand side expression (the function of the observed variables).
  - Use ONLY the observable names above (r, x1, y1, vx1, ..., m1, m2) plus any constants you posit.
  - DO NOT include any 'd.../dt =' or 'a =' prefix.
  - DO NOT include any '=' sign anywhere.
  - Example CORRECT: 'G*r - alpha/r**2'
  - Example WRONG:   'a = G*r - alpha/r**2'
  - Use plain ASCII math: + - * / ** ( ). Constant names may be single Greek-name words (alpha, beta, gamma, kappa) or short variables (G, s1, s2, a, b, c, k). NO LaTeX, NO unicode, NO function calls other than sin/cos/exp/log/sqrt.

Recent Agora context:
<<< ITERATIVE DISCOVERY FEEDBACK — deterministic fit on the OBSERVED data; refine your law to reduce the leftover >>>
CURRENT BEST LAW (across the Council) explains about 94% of the observed motion. A small but SYSTEMATIC leftover remains (it is not noise — it repeats across the trajectory):
  - a steeper inverse-power term than 1/r**2 would explain part of what is left. Its exponent need NOT be a whole number: do not just try 1/r**3 or 1/r**4 — treat the power as a CONTINUOUS value and fit it to the leftover (a non-integer exponent is allowed and likely).
  - an OSCILLATORY component is also present: the leftover changes sign across the distance range (a periodic term, not a pure power).
Propose a symbolic_form that ALSO accounts for this leftover, using only the allowed observable names and any constants you posit.
NOVELTY CHECK: one law-shape has now been proposed 18 times across 38 proposals — the Council is circling the same idea. Do NOT re-propose it; the systematic leftover above is what a genuinely NEW shape must capture.
SEARCH SO FAR (this run): 8 distinct law-shape(s) explored across 38 proposals; the search has begun probing richer multi-term / steeper structure. Build on that history — do not restart from the obvious law.
<<< END FEEDBACK >>>

<<< CONTROLLED-EXPERIMENT FEEDBACK — deterministic fit of YOUR proposed correction to cleaner requested data; commit to the fitted form >>>
CONTROLLED EXPERIMENT RESULT. You requested simpler motion to isolate your proposed correction. On a controlled subset of the data where the motion is cleaner, your proposed correction term beyond the inverse-square force best fits at an inverse-power exponent of about 3.4 (a 1/r**3.4 term), and at that exponent it removes about 100% of what the inverse-square force alone leaves over.
Consider committing to that form: an inverse-square term plus a 1/r**3.4 correction, using only the allowed observable names and any constants you posit. Then refine further if a systematic leftover remains.
<<< END EXPERIMENT FEEDBACK >>>

(no recent Agora context)

Role-specific emphasis:
Demand replication before novelty is accepted. When you see a new hypothesis prefer to write a REFINES edge asking 'did anyone replicate this?' rather than propose a competing hypothesis. Concentrate posterior mass on candidates with at least 2 supporting Agora nodes.

Emit a CouncilHypothesis JSON object with:
  - content: a HypothesisContent (Phase 5.0 schema).
  - posterior: dict[str, float] — mass over the 5
    refinement candidates; sum to 1.0 ± 1e-3, non-negative.
  - candidates_considered: list[str] of length <= 5.
  - reasoning_trace: optional, <= 500 chars.

Consider up to 5 candidates; concentrate mass on
candidates with supporting evidence. Emit the JSON object literally — no markdown, no commentary.
\end{lstlisting}

\end{document}